\documentclass[10pt,twocolumn,letterpaper]{article}

\usepackage{cvpr}              

\usepackage{graphicx}
\usepackage{amsmath}
\usepackage{amssymb}
\usepackage{booktabs}

\usepackage[utf8]{inputenc} 
\usepackage[T1]{fontenc}    
\usepackage{url}            
\usepackage{booktabs}       
\usepackage{amsfonts}       
\usepackage{nicefrac}       
\usepackage{microtype}      
\usepackage{xcolor}         

\usepackage{algorithm}
\usepackage{algorithmicx}
\usepackage{algpseudocode}
\usepackage{graphicx}
\usepackage{amsmath,amssymb}
\usepackage{verbatim}
\usepackage{multirow}
\usepackage{marvosym}
\usepackage{makecell}
\usepackage{bm}
\usepackage{wrapfig}
\usepackage{caption}
\usepackage{accents}
\usepackage{mathtools}
\usepackage{setspace}

\usepackage[accsupp]{axessibility}  

\newcommand{\R}{\ensuremath{\mathbb{R}}}

\def\y{\mathbf{y}}

\renewcommand{\algorithmicrequire}{ \textbf{Input:}} 
\renewcommand{\algorithmicensure}{ \textbf{Output:}}

\usepackage[pagebackref,breaklinks,colorlinks]{hyperref}

\usepackage[capitalize]{cleveref}
\crefname{section}{Sec.}{Secs.}
\Crefname{section}{Section}{Sections}
\Crefname{table}{Table}{Tables}
\crefname{table}{Tab.}{Tabs.}

\def\cvprPaperID{4073} 
\def\confName{CVPR}
\def\confYear{2023}

\begin{document}

\title{Leapfrog Diffusion Model for Stochastic Trajectory Prediction}





\author{
Weibo Mao\textsuperscript{1}, Chenxin Xu\textsuperscript{1}, Qi Zhu\textsuperscript{1}, Siheng Chen\textsuperscript{1,2\footnotemark[1]}, Yanfeng Wang\textsuperscript{2,1},
\\\textsuperscript{1}Shanghai Jiao Tong University,  \textsuperscript{2}Shanghai AI Laboratory
\\
{\tt\small \{kirino.mao,xcxwakaka,georgezhu,sihengc,wangyanfeng\}@sjtu.edu.cn}
}

\maketitle
\renewcommand{\thefootnote}{\fnsymbol{footnote}}
\footnotetext[1]{Corresponding author.}

\begin{abstract}
To model the indeterminacy of human behaviors, stochastic trajectory prediction requires a sophisticated multi-modal distribution of future trajectories. Emerging diffusion models have revealed their tremendous representation capacities in numerous generation tasks, showing potential for stochastic trajectory prediction. However, expensive time consumption prevents diffusion models from real-time prediction, since a large number of denoising steps are required to assure sufficient representation ability. To resolve the dilemma, we present LEapfrog Diffusion model (LED), a novel diffusion-based trajectory prediction model, which provides  real-time, precise, and diverse predictions. The core of the proposed LED is to leverage a trainable leapfrog initializer to directly learn an expressive multi-modal distribution of future trajectories, which skips a large number of denoising steps, significantly accelerating inference speed. Moreover, the leapfrog initializer is trained to appropriately allocate correlated samples to provide a diversity of predicted future trajectories, significantly improving prediction performances. Extensive experiments on four real-world datasets, including NBA/NFL/SDD/ETH-UCY, show that LED consistently improves performance and achieves 23.7\%/21.9\% ADE/FDE improvement on NFL. The proposed LED also speeds up the inference 19.3/30.8/24.3/25.1 times compared to the standard diffusion model on NBA/NFL/SDD/ETH-UCY, satisfying real-time inference needs. Code is available at~\url{https://github.com/MediaBrain-SJTU/LED}.
\end{abstract}

\section{Introduction}


Trajectory prediction aims to predict the future trajectories for one or multiple interacting agents conditioned on their past movements. This task plays a significant role in numerous applications, such as autonomous driving~\cite{levinson2011towards, chen20203d}, drones~\cite{floreano2015science}, surveillance systems~\cite{valera2005intelligent}, human-robot interaction systems~\cite{cheng2020towards}, and interactive robotics~\cite{kanda2002development, li2021symbiotic}. Recently, lots of fascinating research progresses have been made from many aspects, including temporal encoding~\cite{chung2014empirical, vaswani2017attention, giuliari2021transformer, yuan2021agentformer}, interaction modeling~\cite{alahi2016social, gupta2018social, hu2020collaborative, tang2021collaborative, chenxin2022groupnet}, and rasterized prediction~\cite{wu2020motionnet, gilles2021home, mangalam2021goals, gilles2022gohome, zhong2022aware}. In practice, to capture multiple possibilities of future trajectories, a real-world prediction system needs to produce multiple future trajectories. This~leads to the emergence of stochastic trajectory prediction, aiming to precisely model the distribution of future trajectories.

\begin{figure}[t]
  \centering
   \includegraphics[width=\linewidth]{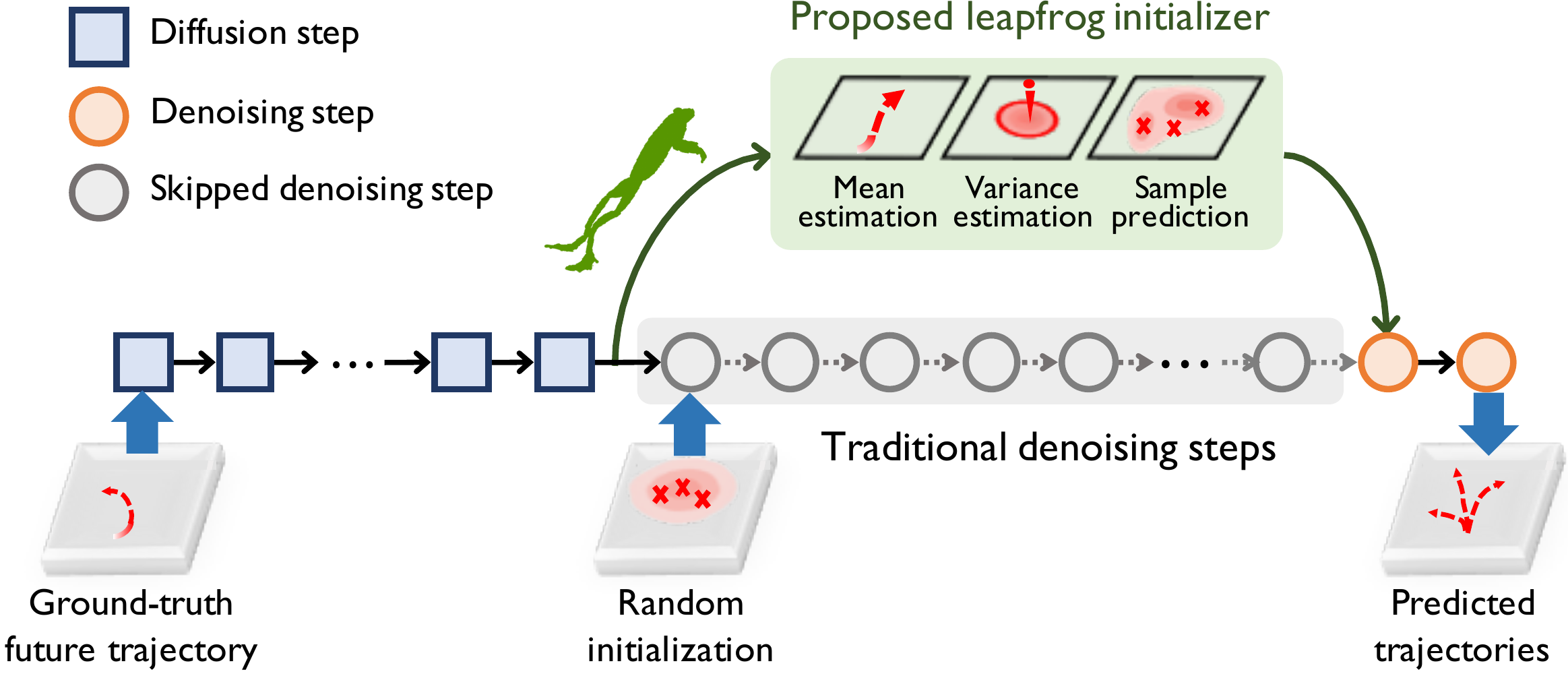}
   \vspace{-5.5mm}
   \caption{Leapfrog diffusion model uses the leapfrog initializer to estimate the denoised distribution and substitute a long sequence of traditional denoising steps, accelerating inference and maintaining representation capacity.}
   \vspace{-4.5mm}
   \label{fig:onecol}
\end{figure}
Previous works have proposed a series of deep generative models for stochastic trajectory prediction. For example,~\cite{gupta2018social, hu2020collaborative} exploit the generator adversarial networks (GANs) to model the future trajectory distribution;~\cite{mangalam2020pecnet, salzmann2020trajectron++, chenxin2022groupnet} consider the conditional variational auto-encoders (CVAEs) structure; and~\cite{apratim2019cfvae} uses the conditional normalizing flow to relax the Gaussian prior in CVAEs and learn more representative priors. Recently, with the great success in image generation~\cite{jonathan2020ddpm, alexander2021iddpm} and audio synthesis~\cite{nanxin2021wavegrad, zhifeng2021diffwave}, denoising diffusion probabilistic models have been applied to time-series analysis and trajectory prediction, and show promising prediction performances~\cite{yusuke2021csdi, tianpei2022mid}. Compared to many other generative models, diffusion models have advantages in stable training and modeling sophisticated distributions through sufficient denoising steps~\cite{pd2021diffusionbeatgans}.

However, there are two critical problems in diffusion models for stochastic trajectory prediction. First,  the real-time inference is time-consuming~\cite {tianpei2022mid}. To ensure the representation ability and generate high-quality samples, an adequate number of denoising steps are required in standard diffusion models, which costs more computational time. For example, experiments show that on the NBA dataset, diffusion models need about $100$ denoising steps to achieve decent prediction performances, which would take $\sim$886ms to predict; while the next frame comes every 200ms. Second, as mentioned in~\cite{inhwan2022npsn}, a limited number of independent and identically distributed samples might not be able to capture sufficient modalities in the underlying distribution of a generative model. Empirically, a few independent sampled trajectories could miss some important future possibilities due to the lack of appropriate sample allocation, significantly deteriorating prediction performances.


In this work, we propose leapfrog diffusion model (LED), a novel diffusion-based trajectory prediction model, which significantly accelerates the inference speed and enables adaptive and appropriate allocations of multiple correlated predictions, providing sufficient diversity in predictions. The core idea of the proposed LED is to learn a rough, yet sufficiently expressive distribution to initialize denoised future trajectories; instead of using a plain Gaussian distribution as in standard diffusion models. Specifically, our forward diffusion process is the same as standard diffusion models, which assures that the ultimate representation ability is pristine; while in the reverse denoising process, we leverage a powerful initializer to produce correlated diverse samples and leapfrog or skip a large number of denoising steps; and then, use only a few denoising steps to refine the distribution. 

To implement such a leapfrog initializer, we consider a reparameterization to alleviate the learning burden. We disassemble a denoised distribution into three parts: mean trajectory, variance, and sample positions under the normalized distribution. To estimate these three, we design three corresponding trainable modules, each of which leverages both a social encoder and a temporal encoder to learn the social-temporal features and produce accurate estimation. Furthermore, all the sample positions are simultaneously generated based on the same social-temporal features, enabling appropriate sample allocations to 
provide diversity.

To evaluate the effectiveness of the proposed method, we conduct experiments on four trajectory prediction datasets: NBA, NFL Football Dataset, Standford Drones Dataset, and ETH-UCY. The quantitative results show we outperform the previous methods and achieve state-of-the-art performance. Specifically, compared to MID~\cite{tianpei2022mid}, the proposed leapfrog diffusion model reduces the average prediction time from $\sim$886ms to $\sim$46ms on the NBA dataset, while achieving a 15.6\%/13.4\% ADE/FDE improvement.

The main contributions are concluded as follows,


    

    $\bullet$ We propose a novel LEapfrog Diffusion model (LED), which is a denoising-diffusion-based stochastic trajectory prediction model. It achieves precise and diverse predictions with fast inference speed. 

    $\bullet$ We propose a novel trainable leapfrog initializer to directly model sophisticated denoised distributions, accelerating inference speed, and adaptively allocating the sample diversity, improving prediction performance.

    $\bullet$ We conduct extensive experiments on four datasets including NBA, NFL, SDD, and ETH-UCY. Results show that i) our approach consistently achieves state-of-the-art performance on all datasets; and ii) our method speeds up the inference by around 20 times compared to the standard diffusion model, satisfying real-time prediction needs.

\section{Related Work}
\label{related_work}
\textbf{Trajectory prediction.} Early works on trajectory prediction focus on a deterministic approach by exploring force models \cite{helbing1995social, MehranOS09}, RNNs\cite{alahi2016social, morton2016analysis, vemula2018social}, and frequency analysis\cite{mao2019learning, mao2020history}. For example, \cite{helbing1995social} models an agent's behavior with attractive and repulsive forces and builds the force equations for prediction. To capture the multi-modalities and model future distribution, recent works start to focus on stochastic trajectory prediction and have proposed a series of deep generative models. Generative Adversarial Network (GAN) structures \cite{gupta2018social, sadeghian2019sophie, SunZH20, FangJSZ20TPNet, hu2020collaborative, dendorfer2021mg} are proposed to generate multiple future trajectory distribution. \cite{mangalam2020pecnet, salzmann2020trajectron++, chenxin2022groupnet, lee2022muse,xu2022dynamic} use the Variational Auto-Encoder (VAE) structure and learn the distribution through variational inference. \cite{apratim2019cfvae} relaxes the Gaussian prior and proposes to use the normalizing flow.  Heatmap  \cite{gilles2021home, mangalam2021goals, gilles2022gohome} is used for modeling future trajectories' distribution on rasterized images. In this work, we propose a new diffusion-based model for trajectory prediction. Compared to previous generative models, our method has a large representation capacity and can model sophisticated trajectory distributions by using a number of diffusion steps. We also enable the correlation between samples to adaptively adjust sample diversity, improving prediction performance.

\textbf{Denoising diffusion probabilistic models.} Denoising diffusion probabilistic models (diffusion models)\cite{jasccha2015dpm, yang2019eg, jonathan2020ddpm} have recently achieved significant results in image generation \cite{pd2021diffusionbeatgans, alexander2021iddpm} and audio synthesis \cite{nanxin2021wavegrad, zhifeng2021diffwave}. The idea of diffusion models is first proposed by DPM~\cite{jasccha2015dpm}, which imitates the diffusion process in non-equilibrium statistical physics and reconstructs the data distribution using the denoising model. Later, \cite{kashif2021timegrad, yusuke2021csdi} propose diffusion models, combining with the seq-to-seq models, for probabilistic time series forecasting. MID~\cite{tianpei2022mid} is the first to build diffusion models for trajectory prediction in modeling the indeterminacy variation process. 

The standard diffusion models use hundreds of denoising steps, preventing these models from real-time applications. To accelerate the sampling process, DDIM~\cite{jiaming2021ddim} first predicts the original data and then estimates the direction to the next expected timestamp based on the non-Markov process. PD~\cite{Salimans2022pg} applies the knowledge distillation on the denoising steps with a deterministic diffusion sampler, which will be repeated for times to accelerate the sampling. All these fast sampling methods start denoising from noise inputs, which are randomly and independently initialized.  
In this work, we use a trainable leapfrog initializer to initialize a sufficiently expressive distribution, which replaces a large number of former denoising steps for much faster inference speed. 
%


\begin{figure*}[t]
    \vspace{-10mm}
  \centering
   \includegraphics[width=0.95\linewidth]{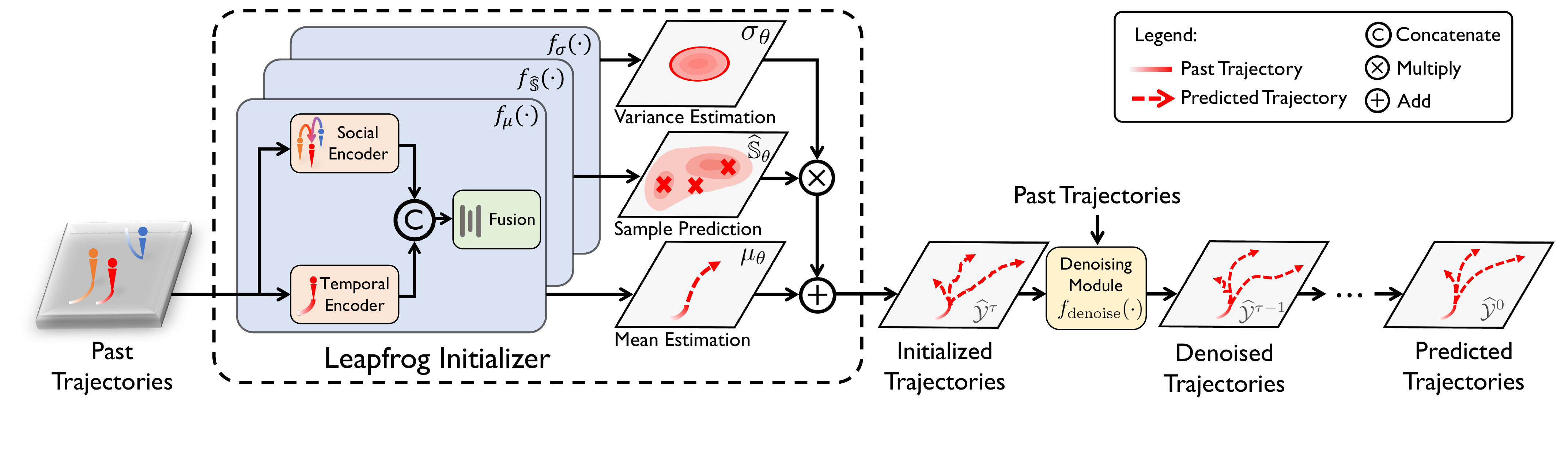}
   \vspace{-8.5mm}
   \caption{\small Proposed leapfrog diffusion model (LED) in inference phase. The red agent is the to-be-predicted agent. LED first predicts $K$ initialized trajectories at $\tau$th denoised step through a trainable leapfrog initializer. Then, followed by a few denoising steps, LED obtains the final predictions.
   In leapfrog initializer, LED learns statistics and generates correlated samples with the reparameterization.
   }
   \vspace{-4.5mm}
   \label{fig:design}
\end{figure*}


\section{Background}
\subsection{Problem Formulation}
Trajectory prediction aims to predict an agent's future trajectory based on the past trajectories of itself and surrounding agents.  For a to-be-predicted agent, let $\mathbf{X}=[\mathbf{x}^{-T_{\rm p}+1}, \mathbf{x}^{-T_{\rm p}+2}, \dots, \mathbf{x}^{0}]\in \mathbb{R}^{T_{\rm p} \times 2}$ be the observed past trajectory over $T_{\rm p}$ timestamps where $\mathbf{x}^t\in \mathbb{R}^2$ records the 2D spatial coordinate at timestamp $t$. Let $\mathcal{N}$ be the neighbouring agent set and $\mathbb{X}_{\mathcal{N}} = [\mathbf{X}_{\mathcal{N}_1} ,\mathbf{X}_{\mathcal{N}_2},\cdots,\mathbf{X}_{\mathcal{N}_L}] \in \R^{L \times T_{\rm p} \times 2}$ be the past trajectories of neighbours, where $\mathbf{X}_{\mathcal{N}_\ell}\in \R^{T_{\rm p} \times 2}$ is the trajectory of the $\ell$th neighbour. The corresponding ground-truth future trajectory for the to-be-predicted agent is $\mathbf{Y}=[\mathbf{y}^{1}, \mathbf{y}^{2}, \dots, \mathbf{y}^{T_{\rm f}}]\in \mathbb{R}^{T_{\rm f} \times 2}$ over $T_{\rm f}$ timestamps, where $\y^t \in \R^2$ is the 2D coordinate at future timestamp $t$.

Because of the indeterminacy of future trajectories, it is usually more reliable to predict more than one trajectory to capture multiple possibilities. Here we consider stochastic trajectory prediction, which predicts the distribution of a future trajectory, instead of a single future trajectory. The goal of stochastic trajectory prediction is to train a prediction model $g_\theta(\cdot)$ with parameters $\theta$ to generate a distribution
$\mathcal{P}_\theta = g_\theta(\mathbf{X}, \mathbb{X}_{\mathcal{N}})$. Based on this distribution $\mathcal{P}_\theta$, we can draw $K$ samples, $\widehat{\mathcal{Y}} = \{\widehat{\mathbf{Y}}_1, \widehat{\mathbf{Y}}_2, \dots, \widehat{\mathbf{Y}}_K \}$, so that at least one sample is close to the ground-truth future trajectory. The overall learning problem is
\begin{equation}
\setlength{\abovedisplayskip}{4pt}
   \setlength{\belowdisplayskip}{2pt}
  \theta^* = \min_{\theta} \min_{\widehat{\mathbf{Y}}_i \in \widehat{\mathcal{Y}}} D(\widehat{\mathbf{Y}}_i, \mathbf{Y}), ~~~{\rm s.t.}~~~ \widehat{\mathcal{Y}} \sim \mathcal{P}_\theta.
\end{equation}

\subsection{Diffusion Model for Trajectory Prediction}
Here we present a standard diffusion model for trajectory prediction, which lays a foundation for the proposed method. The core idea is to learn and refine a sophisticated underlying distribution of trajectories through cascading a series of simple denoising steps. To implement this, a diffusion model performs a forward diffusion process to intentionally add a series of noises to a ground-truth future trajectory; and then, it uses a conditional denoising process to recover the future trajectory from noise inputs conditioned on past trajectories.

Mathematically, let $\mathbf{X}$ and $\mathbb{X}_{\mathcal{N}}$ be the past trajectories of the ego agent and the neighboring agents, respectively, and $\mathbf{Y}$ be the future trajectory of the ego agent. The diffusion model for trajectory prediction works as follows,
\begin{subequations}
\setlength{\abovedisplayskip}{2pt}
   \setlength{\belowdisplayskip}{2pt}
\label{eq:diffusion_model}
    \begin{align}
        \label{eq:initialization_step}
        &\mathbf{Y}^{0}  =  \mathbf{Y},
        \\ 
        \label{eq:diffusion_step}
        &\mathbf{Y}^{\gamma}  =  f_{\operatorname{diffuse}}(\mathbf{Y}^{\gamma-1}),  \ \gamma=1, \cdots, \Gamma,
        \\
        \label{eq:sampling_step}
        &\widehat{\mathbf{Y}}^\Gamma_k  \stackrel{i.i.d}{\sim}   \mathcal{P}(\widehat{\mathbf{Y}}^\Gamma)=\mathcal{N}(\widehat{\mathbf{Y}}^\Gamma; \textbf{0}, \mathbf{I}), {\rm sample}~K~\operatorname{times},
        \\
        \label{eq:denoising_step}
        &\widehat{\mathbf{Y}}^{\gamma}_k  =  f_{\operatorname{denoise}}(\widehat{\mathbf{Y}}^{\gamma+1}_k, \mathbf{X}, \mathbb{X}_{\mathcal{N}}),  \;\gamma\!=\!\Gamma\!-\!1,\!\cdots\!,\!0,
    \end{align}
\end{subequations}
where $\mathbf{Y}^{\gamma}$ is the noisy trajectory at the $\gamma$th diffusion step and $\widehat{\mathbf{Y}}^{\gamma}_k$ is the $k$th sample of denoised trajectory at the $\gamma$th denoising step. The final $K$ predicted trajectories are $\widehat{\mathcal{Y}} = \{\widehat{\mathbf{Y}}^0_1, \widehat{\mathbf{Y}}^0_2, \dots, \widehat{\mathbf{Y}}^0_K \}.$

Step~\eqref{eq:initialization_step} initializes the diffused trajectory; Step~\eqref{eq:diffusion_step} uses a forward diffusion operation $f_{\operatorname{diffuse}}(\cdot)$ to successively add noises to $\mathbf{Y}^{\gamma-1}$ and obtain the diffused trajectory $\mathbf{Y}^{\gamma}$; Step~\eqref{eq:sampling_step} draws $K$ independent and identically distributed samples to initialize denoised trajectories $\widehat{\mathbf{Y}}^\Gamma_k$ from a normal distribution; and Step~\eqref{eq:denoising_step} iteratively applies a denoising operation $f_{\operatorname{denoise}}(\cdot)$ to obtain the denoised  trajectory $\widehat{\mathbf{Y}}^{\gamma}_k$ conditioned on past trajectories $\mathbf{X}, \mathbb{X}_{\mathcal{N}}$.  Note that i) Steps~\eqref{eq:initialization_step} and~\eqref{eq:diffusion_step} correspond to the forward diffusion process and are not used in inference; ii) During training, $\mathbf{Y}^{\gamma}$ is naturally the supervision for $\widehat{\mathbf{Y}}^\gamma_k$ at the $\gamma$th step. Conceptually, each denoising step is the reverse of the diffusion step, and each pair of $\mathbf{Y}^{\gamma}$ and $\widehat{\mathbf{Y}}^{\gamma}_k$ shares the same underlying distribution.

The standard diffusion model is expressively powerful in learning sophisticated distributions and has achieved great success in many generation tasks. However, the task of motion prediction requires real-time inference but the running time of a diffusion model is constrained by the large number of denoising steps. Meanwhile, less denoising steps usually cause a weaker representation ability of future distributions. 
To achieve higher efficiency while preserving a promising representation ability, we propose leapfrog diffusion model, which uses a trainable initializer to capture sophisticated distributions and substitute a large number of denoising steps.

\section{Leapfrog Diffusion Model}

\subsection{System Architecture}
In this section, we propose the leapfrog diffusion model. Here leapfrog means that a large number of small denoising steps can be replaced by a single, yet powerful leapfrog initializer, which can significantly accelerate the inference speed without losing representation ability. Let $\mathbf{X}$ and $\mathbb{X}_{\mathcal{N}}$ be the past trajectories of the ego agent and its neighboring agents, and $\mathbf{Y}$ be the future trajectory of the ego agent. Denote $\tau$ as the leapfrog step. The overall procedure of the proposed leapfrog diffusion model is formulated as follows,
\begin{subequations}
  \setlength{\abovedisplayskip}{-5pt}
   \setlength{\belowdisplayskip}{2pt}
\label{eq:leapfrog_diffusion_model}
    \begin{align} \label{eq:leapfrog_initialization_step}& \mathbf{Y}^{0}  =  \mathbf{Y},
        \\ 
        \label{eq:leapfrog_diffusion_step}
        & \mathbf{Y}^{\gamma}  =  f_{\operatorname{diffuse}}(\mathbf{Y}^{\gamma-1}),  \ \gamma=1, \cdots, \Gamma,
        \\
        \label{eq:leapfrog_isg}
       & \widehat{\mathcal{Y}}^\tau  \stackrel{K}{\sim}   \mathcal{P}(\widehat{\mathbf{Y}}^\tau) = f_{\operatorname{LSG}}(\mathbf{X}, \mathbb{X}_{\mathcal{N}}),
        \\
        \label{eq:leapfrog_denoising_step}
       & \widehat{\mathbf{Y}}^{\gamma}_k  =  f_{\operatorname{denoise}}(\widehat{\mathbf{Y}}^{\gamma+1}_k, \mathbf{X}, \mathbb{X}_{\mathcal{N}}),  \;\gamma=\tau\!-\!1, \cdots, 0. 
    \end{align}
\end{subequations}

Compared to the standard diffusion model~\eqref{eq:diffusion_model}, the main difference lies in Step~\eqref{eq:leapfrog_isg}.  The standard diffusion initializes the $\Gamma$th denoised distribution $\mathcal{P}(\widehat{\mathbf{Y}}^\Gamma)$ by a plain normal distribution~\eqref{eq:sampling_step} and requires a lot of denoising steps to enrich the expressiveness of the denoised distribution; while in Step~\eqref{eq:leapfrog_isg}, we propose a novel leapfrog initializer $f_{\operatorname{LSG}}(\cdot)$ to directly model the $\tau$th denoised distribution $\mathcal{P}(\widehat{\mathbf{Y}}^\tau)$, which is hypothetically equivalent to the output of executing $(\Gamma-\tau)$ denoising steps~\eqref{eq:denoising_step}. We then draw samples from the distribution $\mathcal{P}(\widehat{\mathbf{Y}}^\tau)$ and obtain $K$ future trajectories $\widehat{\mathcal{Y}}^\tau = \{\widehat{\mathbf{Y}}^\tau_1, \widehat{\mathbf{Y}}^\tau_2, \dots, \widehat{\mathbf{Y}}^\tau_K \},$ where $\stackrel{K}{\sim}$ in \eqref{eq:leapfrog_denoising_step} means $K$ samples are dependent to intentionally allocate appropriate sample diversity. Then, in Step~\eqref{eq:leapfrog_denoising_step}, we only need to apply the remaining $\tau$ denoising steps for each trajectory $\widehat{\mathbf{Y}}^\gamma_k$ to obtain the final prediction $\widehat{\mathcal{Y}} = \{\widehat{\mathbf{Y}}^0_1, \widehat{\mathbf{Y}}^0_2, \dots, \widehat{\mathbf{Y}}^0_K \}$.


Note that i) the proposed leapfrog diffusion model reduces the denoising steps from $\Gamma$ to $\tau(\ll\!\Gamma)$ in Step~\eqref{eq:leapfrog_denoising_step} as the leapfrog initializer directly provides the trajectories at denoising step $\tau$, accelerating the inference; ii) instead of taking independent and identically distributed samples in Step~\eqref{eq:sampling_step}, the proposed leapfrog initializer generates $K$ trajectories $\widehat{\mathcal{Y}}^\tau$  simultaneously in Step~\eqref{eq:leapfrog_isg}, allowing $K$ samples to be aware of each other; 
and iii) the standard diffusion model and the proposed leapfrog diffusion model share the same forward diffusion process, assuring that the representation capacity is not reduced.

\subsection{Leapfrog Initializer}
We now dive into the design details of the proposed leapfrog initializer, which leapfrogs $(\Gamma-\tau)$ denoising steps. In leapfrog initializer, we model the $\tau$th denoised distribution $\mathcal{P}(\widehat{\mathbf{Y}}^\tau)$ through learning models. However, it is nontrivial for a learning model to directly capture the sophisticated distribution, which usually causes unstable training. To ease the learning burden of the model, we disassemble the distribution $\mathcal{P}(\widehat{\mathbf{Y}}^\tau)$ into three representative parts: the mean, global variance and sample prediction. For each part, we design trainable modules correspondingly. Mathematically, let $\mathbf{X}$ and $\mathbb{X}_{\mathcal{N}}$ be the past trajectories of the ego agent and the neighboring agents, respectively. The proposed leapfrog initializer generates $K$ samples as follows, 
\begin{eqnarray}
\setlength{\abovedisplayskip}{2pt}
   \setlength{\belowdisplayskip}{2pt}
    && \mu_\theta = f_\mu (\mathbf{X}, \mathbb{X}_{\mathcal{N}}) \in  \mathbb{R}^{T_{\rm f} \times 2},  \quad
    \nonumber
    \\ \nonumber
    &&  \sigma_\theta = f_\sigma (\mathbf{X}, \mathbb{X}_{\mathcal{N}})  \in  \mathbb{R},
    \\ \nonumber
    &&  \widehat{\mathbb{S}}_{\theta} = [\widehat{\mathbf{S}}_{\theta, 1}, \cdots, \widehat{\mathbf{S}}_{\theta, K}]  = f_{\widehat{\mathbb{S}}}(\mathbf{X}, \mathbb{X}_{\mathcal{N}}, \sigma_\theta ) \in \mathbb{R}^{T_{\rm f} \times 2  \times K},
    \\ \label{eq:sample_reparameterization}
    && \widehat{\mathbf{Y}}^\tau_k = \mu_\theta + \sigma_\theta \cdot \widehat{\mathbf{S}}_{\theta, k} \in \mathbb{R}^{T_{\rm f} \times 2},
\end{eqnarray}
where $f_\mu(\cdot), f_\sigma(\cdot), f_{\widehat{\mathbb{S}}}(\cdot)$ are three trainable modules, $\mu_\theta, \sigma_\theta$ are the mean and standard deviation of $\mathcal{P}(\widehat{\mathbf{Y}}^\tau)$, respectively, and $\widehat{\mathbf{S}}_{\theta, k}$ is the normalized positions for the $k$th sample. 

To be specific, the mean estimate module $f_\mu (\cdot)$ infers the mean trajectory of the $\tau$th denoised distribution $\widehat{\mathcal{P}}(\widehat{\mathbf{Y}}^\tau)$ with past trajectories $(\mathbf{X}, \mathbb{X}_{\mathcal{N}})$. The mean trajectory $\mu_\theta$ is shared across all the $K$ samples. The variance estimate module $f_\sigma (\cdot)$ infers the standard deviation of the $\tau$th denoised distribution $\widehat{\mathcal{P}}(\widehat{\mathbf{Y}}^\tau)$, reflecting the overall uncertainty of the trajectory, which is also shared across all the $K$ samples.  The sample prediction module $f_{\widehat{\mathbb{S}}}(\cdot)$ takes the past trajectories $(\mathbf{X}, \mathbb{X}_{\mathcal{N}})$ and the predicted uncertainty $\sigma_\theta $ as the input and predicts $K$ normalized positions where each $\widehat{\mathbf{S}}_{\theta, k} \in \mathbb{R}^{T_{\rm f} \times 2}$. 

Note that i) the reparameterization in Eq.~\eqref{eq:sample_reparameterization} allows us to avoid learning a raw sophisticated distribution, making the training much easier; and ii) $K$ normalized predictions are generated simultaneously from the same underlying feature, assuring appropriately allocated trajectories with variance estimation and better capturing the multi-modalities.

To implement the three trainable modules: $f_\mu(\cdot)$, $f_\sigma(\cdot)$, $f_{\widehat{\mathbb{S}}}(\cdot)$, we consider a similar network design: a social encoder to capture social influence, a temporal encoder to learn temporal embedding, and an aggregation layer to fuse both social and temporal information; see Figure \ref{fig:design}. Here we take the mean estimation module $f_\mu(\cdot)$ as an example. The mean trajectory is obtained as follows,
\begin{subequations}
\begin{align}
        \label{eq:mean_social}
        &\mathbf{e}^{\operatorname{social}}_{\mu_\theta} = \operatorname{softmax}\Big(\dfrac{f_\mathrm{q} (\mathbf{X})f_\mathrm{k}(\mathbb{X}_\mathcal{N})^{\mathsf{T}}}{\sqrt{d}}\Big) f_\mathrm{v}(\mathbb{X}_\mathcal{N}),
        \\
        \label{eq:mean_temporal}
        &\mathbf{e}^{\operatorname{temp}}_{\mu_\theta}  =  f_{\operatorname{GRU}} (f_{\operatorname{conv1D}}(\mathbf{X})),
        \\
        \label{eq:mean_fusion}
        &\mu_\theta  =  f_{\operatorname{fusion}}([\mathbf{e}^{\operatorname{social}}_{\mu_\theta} : \mathbf{e}^{\operatorname{temp}}_{\mu_\theta}]).
\end{align}
\end{subequations}
Step~\eqref{eq:mean_social} obtains the social embedding $\mathbf{e}^{\operatorname{social}}_{\mu_\theta}$ based on the multi-head attention with $d$ the embedding dimension and $f_\mathrm{q}(\cdot), f_\mathrm{k}(\cdot), f_\mathrm{v}(\cdot)$ the query/key/value embedding functions. Step~\eqref{eq:mean_temporal} obtains the temporal embedding through the feature encoder $f_{\operatorname{conv1D}}(\cdot)$, mapping the raw coordinates into the high-dimensional feature, followed by the gated recurrent units $f_{\operatorname{GRU}}(\cdot)$, capturing the temporal dependence in the high dimensional sequence. Step~\eqref{eq:mean_fusion} concatenates both social and temporal embeddings and 
uses a multi-layer perceptron $f_{\operatorname{fusion}}(\cdot)$ to obtain the final mean estimation. 
Note that the sample prediction module $f_{\widehat{\mathbb{S}}}(\cdot)$ also takes the estimated standard deviation as the input, working as
\begin{equation}
\nonumber
\label{eq:LSG_sampling}
    \begin{aligned}
&\mathbf{e}^{\operatorname{\sigma}}_{\widehat{\mathbb{S}}_{\theta}}  = f_{\operatorname{encode}} (\sigma_\theta),
        \\
       & \widehat{\mathbb{S}}_{\theta} = f_{\operatorname{fusion}}([\mathbf{e}^{\operatorname{social}}_{\widehat{\mathbb{S}}_{\theta}} : \mathbf{e}^{\operatorname{temp}}_{\widehat{\mathbb{S}}_{\theta}} : \mathbf{e}^{\sigma}_{\widehat{\mathbb{S}}_{\theta}}]),
    \end{aligned}
\end{equation}
where an encoder $f_{\operatorname{encode}}(\cdot)$ operates on the estimated variance $\sigma_\theta$ and generates high dimensional embedding $\mathbf{e}^{\operatorname{\sigma}}_{\widehat{\mathbb{S}}_{\theta}}$. By this, the variance estimation also involves in the sample prediction process, instead of just scaling these prediction.

After obtaining $K$ samples $\widehat{\mathcal{Y}}^\tau = \{\widehat{\mathbf{Y}}^\tau_1, \widehat{\mathbf{Y}}^\tau_2, \dots, \widehat{\mathbf{Y}}^\tau_K \}$ from leapfrog initializer, we execute the remaining $\tau$ denoising steps to iteratively refine those predicted trajectories~\eqref{eq:leapfrog_denoising_step}.

\subsection{Denoising Module}
Here we elaborate the design of a denoising module $f_{\operatorname{denoise}}(\cdot)$, which denoises the trajectory $\widehat{\mathbf{Y}}_k^{\gamma+1}$ conditioned on past trajectories $(\mathbf{X}, \mathbb{X}_{\mathcal{N}})$. In a denoising module, two parts are trainable: a transformer-based context encoder $f_{\operatorname{context}}(\cdot)$ to learn a social-temporal embedding and a noise estimation module $f_{\bm{\epsilon}}(\cdot)$ to estimate the noise to reduce. Mathematically, the $\gamma$th denoising step works as follows,
\begin{subequations}
\setlength{\abovedisplayskip}{0pt}
   \setlength{\belowdisplayskip}{4pt}
\begin{align}
        \label{eq:context_encoder}
        &\mathbf{C}  =  f_{\operatorname{context}}(\mathbf{X}, \mathbb{X}_{\mathcal{N}} ),
        \\
        \label{eq:noise_estimate}
        &\bm{\epsilon}_\theta^{\gamma} =  f_{\bm{\epsilon}}(\widehat{\mathbf{Y}}_k^{\gamma+1}, \mathbf{C}, \gamma+1), 
        \\\label{eq:core_denoising}
        &\widehat{\mathbf{Y}}_k^{\gamma} =  \dfrac{1}{\sqrt{\alpha_{\gamma}}}(\widehat{\mathbf{Y}}_k^{\gamma+1} \!\!-\!\! \dfrac{1-\alpha_\gamma}{\sqrt{1-\bar{\alpha}_\gamma}} \bm{\epsilon}_\theta^{\gamma} ) \!+ \!\!\sqrt{1\!-\!\alpha_\gamma}\mathbf{z},
\end{align}
\end{subequations}
where $\alpha_\gamma$ and $\bar{\alpha}_\gamma = \prod_{i=1}^\gamma \alpha_i$ are parameters in the diffusion process and $\mathbf{z}\sim \mathcal{N}(\mathbf{z}; \mathbf{0}, \mathbf{I})$ is a noise. Step~\eqref{eq:context_encoder} uses a context encoder $f_{\operatorname{context}}(\cdot)$ on past trajectories $(\mathbf{X}, \mathbb{X}_{\mathcal{N}})$ to obtain the context condition $\mathbf{C}$, which shares a similar structure to mean estimation module $f_\mu(\cdot)$; Step~\eqref{eq:noise_estimate} estimates the noise $\bm{\epsilon}_\theta^{\gamma}$ in the noisy trajectory $\widehat{\mathbf{Y}}_k^{\gamma+1}$ through noise estimation $f_{\bm{\epsilon}}(\cdot)$ implemented by multi-layer perceptions with the context $\mathbf{C}$; Step~\eqref{eq:core_denoising} provides a standard denoising step~\cite{jonathan2020ddpm}; see more details in the supplementary material.

\subsection{Training Objective}
\label{training_objective}
To train a leapfrog diffusion model, we consider a two-stage training strategy, where the first stage trains a denoising module and  the second stage focuses on a leapfrog initializer. The reason to use two stages is because the training of leapfrog initializer is more stable given fixed distribution $\mathcal{P}(\widehat{\mathbf{Y}}^\tau)$, avoiding non-convergent training.

Concretely, the first stage trains a denoising module $f_{\operatorname{denoise}}(\cdot)$ in Step~\eqref{eq:leapfrog_denoising_step}  based on a standard training schedule of a diffusion model~\cite{jonathan2020ddpm, tianpei2022mid} through noise estimation loss: 
\begin{eqnarray}
\nonumber
\label{eq:loss_diffusion}
    \mathcal{L_{\operatorname{NE}}} = \Vert \bm{\epsilon} - f_{\bm{\epsilon}}(\mathbf{Y}^{\gamma+1}, f_{\operatorname{context}}(\mathbf{X}, \mathbb{X}_{\mathcal{N}} ), \gamma+1)\Vert_2,
\end{eqnarray}
where $\gamma \sim \operatorname{U}\{1, 2, \cdots, \Gamma\}$, $\bm{\epsilon}\sim \mathcal{N}(\bm{\epsilon}; \mathbf{0}, \mathbf{I})$ and the diffused trajectory $\mathbf{Y}^{\gamma+1} = \sqrt{\bar{\alpha}_\gamma}~\mathbf{Y}^0 + \sqrt{1-\bar{\alpha}_\gamma}\bm{\epsilon}$. We then back-propagate this loss and train the parameters in the context encoder $f_{\operatorname{context}}(\cdot)$ and the noise estimation module $f_{\bm{\epsilon}}(\cdot)$.

In the second stage, we optimize a leapfrog diffusion model with a trainable leapfrog initializer and frozen denoising modules. For each sample, the loss function is 
\begin{eqnarray}
\vspace{-10mm}
\nonumber
\label{eq:loss_implicit}
    \mathcal{L} &=& \mathcal{L}_{\operatorname{distance}} + \mathcal{L}_{\operatorname{uncertainty}} 
    \nonumber \\ \nonumber
    &=& w \cdot \min_k  \Vert \mathbf{Y} \!\!-\!\!\widehat{\mathbf{Y}}_k \Vert_2 + \Big( \frac{ \sum_{k} \Vert \mathbf{Y} \!\!-\!\!\widehat{\mathbf{Y}}_k \Vert_2}{\sigma_\theta^2 K} + \log \sigma_\theta^2 \Big),
\vspace{-10mm}
\end{eqnarray}
where $w \in \mathbb{R}$ is a hyperparameter weight. The first term constrains the minimum distance in $K$ predictions. Intuitively, if a leapfrog initializer generates high-quality estimations for distribution $\mathcal{P}(\widehat{\mathbf{Y}}^\tau)$, then one of the $K$ predictions in $\widehat{\mathcal{Y}}$ should be close to the ground-truth trajectory $\mathbf{Y}$. The second term normalizes the variance estimation $\sigma_\theta$ in reparameterization \eqref{eq:sample_reparameterization} through an uncertainty loss, balancing the prediction diversity and mean accuracy. Note that the variance estimation controls the dispersion of the predictions, bridging scenery complexity and prediction diversity. The first part $\frac{ \sum_{k} \Vert \mathbf{Y} \!\!-\!\!\widehat{\mathbf{Y}}_k \Vert_2}{\sigma_\theta^2 K}$ makes the value of $\sigma_\theta$ proportional to the complexity of the scenario. The second part $\log \sigma_\theta^2$ is a regulariser used to avoid a trivial solution for $\sigma_\theta$, i.e., generating high variance for all predictions.

\begin{algorithm}[t]
\setstretch{1.15}
    \renewcommand{\algorithmicrequire}{\textbf{Input:}}
	\renewcommand{\algorithmicensure}{\textbf{Output:}}
	\caption{Leapfrog Diffusion Model in Inference} 
	\begin{algorithmic}[1]
		\Require Observed trajectories $\mathbf{X}, \mathbb{X}_{\mathcal{N}}$, Leapfrog step $\tau$
		\Ensure Predicted trajectories $\widehat{\mathcal{Y}}$
		\State $\mu_\theta = f_\mu(\mathbf{X}, \mathbb{X}_{\mathcal{N}})$ \Comment{Mean estimation}
		\State $\sigma_\theta = f_\sigma(\mathbf{X}, \mathbb{X}_{\mathcal{N}})$ \Comment{Variance estimation}
		\State $\widehat{\mathbb{S}}_{\theta} = f_{\widehat{\mathbb{S}}}(\mathbf{X}, \mathbb{X}_{\mathcal{N}}, \sigma_\theta )$ \Comment{Sample prediction}
		\State $\widehat{\mathbf{Y}}^\tau_k = \mu_\theta \!+\! \sigma_\theta \!\cdot\! \widehat{\mathbf{S}}_{\theta, k}, k=1,\!\cdots\!,K$ \Comment{Reparameterization
} 
		\For{$\gamma=\tau-1, ..., 0$} 
		\State $\widehat{\mathbf{Y}}^{\gamma}_k  =  f_{\operatorname{denoise}}(\widehat{\mathbf{Y}}^{\gamma+1}_k, \mathbf{X}, \mathbb{X}_{\mathcal{N}})$ \Comment{Denoising step}
		\EndFor 
		\State $\widehat{\mathcal{Y}} = \widehat{\mathcal{Y}}^0=\{\widehat{\mathbf{Y}}^{0}_1,\cdots,\widehat{\mathbf{Y}}^{0}_K\}$ \\
		\Return $\widehat{\mathcal{Y}}$
	\end{algorithmic} 
	\label{alg:sampling} 
\end{algorithm}

\begin{table*}[!t]
\vspace{-13pt}
\footnotesize
\centering
\setlength{\tabcolsep}{1mm}{\caption{\small Comparison with baseline models on NBA dataset. minADE$_{20}$ /minFDE$_{20}$ (meters) are reported. \textbf{Bold}/\underline{underlined} fonts represent the best/second-best result. Compared to the previous SOTA method, MID, our method achieves a 15.6\%/13.4\% ADE/FDE improvement.}
\vspace{-10pt}
\fontsize{7.8}{9.8}\selectfont
\begin{tabular}{l|cccccccccc|c}
\hline
\hline
     \multirow{2}{*}{Time} 
    &\makecell[c]{Social-\\GAN\cite{gupta2018social}}
    &\makecell[c]{STGAT\cite{huang2019stgat}}
    & \makecell[c]{Social-\\STGCNN\cite{mohamed2020social}}
    & \makecell[c]{PECNet\cite{mangalam2020pecnet}}
    &\makecell[c]{STAR\cite{yu2020spatio}}
    &\makecell[c]{Trajectron++\\ \cite{salzmann2020trajectron++}} 
    &\makecell[c]{MemoNet \\\cite{chenxin2022memonet}}  
    &\makecell[c]{NPSN \cite{inhwan2022npsn}}
    &\makecell[c]{GroupNet \\\cite{chenxin2022groupnet}}  
    &\makecell[c]{MID \\\cite{tianpei2022mid}}  
    &\multirow{2}{*}{\textbf{Ours}}
    \\
    & \color{blue}{\scriptsize{CVPR'18}}
    & \color{blue}{\scriptsize{ICCV'19}}
    & \color{blue}{\scriptsize{CVPR'20}}
    & \color{blue}{\scriptsize{ECCV'20}}
    & \color{blue}{\scriptsize{ECCV'20}}
    & \color{blue}{\scriptsize{ECCV'20}}
    & \color{blue}{\scriptsize{CVPR'22}}
    & \color{blue}{\scriptsize{CVPR'22}}
    & \color{blue}{\scriptsize{CVPR'22}}
    & \color{blue}{\scriptsize{CVPR'22}}
    &
    \\
\hline
     1.0s & 0.41/0.62 & 0.35/0.51 & 0.34/0.48 & 0.40/0.71 & 0.43/0.66 & 0.30/0.38 & 0.38/0.56 & 0.35/0.58 & \underline{0.26}/\underline{0.34} & 0.28/0.37 & \textbf{0.18}/\textbf{0.27}\\
     2.0s & 0.81/1.32 & 0.73/1.10 & 0.71/0.94 & 0.83/1.61 & 0.75/1.24 & 0.59/0.82 & 0.71/1.14 &  0.68/1.23 & \underline{0.49}/\underline{0.70} & 0.51/0.72 & \textbf{0.37}/\textbf{0.56}\\
     3.0s & 1.19/1.94 & 1.04/1.75 & 1.09/1.77 & 1.27/2.44 & 1.03/1.51 & 0.85/1.24 & 1.00/1.57 & 1.01/1.76 & 0.73/1.02 & \underline{0.71}/\underline{0.98} & \textbf{0.58}/\textbf{0.84}\\
     Total(4.0s) & 1.59/2.41 & 1.40/2.18 & 1.53/2.26 & 1.69/2.95 & 1.13/2.01 &  1.15/1.57 & 1.25/1.47 & 1.31/1.79 & \underline{0.96}/1.30 & \underline{0.96}/\underline{1.27} & \textbf{0.81}/\textbf{1.10}\\
\hline
\end{tabular}
\label{table:nba}}
\vspace{-3mm}
\end{table*}

\begin{table*}[!t]
\footnotesize
\centering
\setlength{\tabcolsep}{1mm}{\caption{\small 
Comparison with baseline models on NFL dataset. minADE$_{20}$/minFDE$_{20}$ (meters) are reported.
\textbf{Bold}/\underline{underlined} fonts represent the best/second-best result. Compared to the previous SOTA method, MID, our method achieves a 23.7\%/21.9\%  improvement.}
\vspace{-10pt}
\fontsize{7.8}{9.8}\selectfont
\begin{tabular}{l|cccccccccc|c}
\hline
\hline
    \multirow{2}{*}{Time} 
    &\makecell[c]{Social-\\GAN\cite{gupta2018social}}
    &\makecell[c]{STGAT\cite{huang2019stgat}}
    &\makecell[c]{Social-\\STGCNN\cite{mohamed2020social}}
    & \makecell[c]{PECNet\cite{mangalam2020pecnet}}
    &\makecell[c]{STAR\cite{yu2020spatio}}
    &\makecell[c]{Trajectron++\\ \cite{salzmann2020trajectron++}} 
    & \makecell[c]{LB-EBM \\ \cite{PangZ0W2021ebm}}
    & \makecell[c]{NPSN\cite{inhwan2022npsn}}
    &\makecell[c]{GroupNet \\ \cite{chenxin2022groupnet}\\}  
    & \makecell[c]{MID\cite{tianpei2022mid}}
    &\multirow{2}{*}{\textbf{Ours}}
    \\
    & \color{blue}{\scriptsize{CVPR'18}} 
    & \color{blue}{\scriptsize{ICCV'19}} 
    & \color{blue}{\scriptsize{CVPR'20}} 
    & \color{blue}{\scriptsize{ECCV'20}} 
    & \color{blue}{\scriptsize{ECCV'20}} 
    & \color{blue}{\scriptsize{ECCV'20}} 
    & \color{blue}{\scriptsize{CVPR'21}} 
    & \color{blue}{\scriptsize{CVPR'22}} 
    & \color{blue}{\scriptsize{CVPR'22}} 
    & \color{blue}{\scriptsize{CVPR'22}} 
    &
    \\
\hline
1.0s            & 0.37/0.68 & 0.35/0.64 & 0.45/0.64 & 0.52/0.97 & 0.49/0.84 & 0.41/0.65 & 0.75/1.05 & 0.43/0.64 & 0.32/\underline{0.57} & \underline{0.30}/0.58 & \textbf{0.21}/\textbf{0.34} \\
2.0s            & 0.83/1.53 & 0.82/1.60 & 1.06/1.87 & 1.19/2.47 & 1.02/1.84 & 0.93/1.65 & 1.26/2.28 & 0.83/1.52 & 0.73/1.39 & \underline{0.71}/\underline{1.31} & \textbf{0.49}/\textbf{0.91} \\
Total(3.2s)     & 1.44/2.51 & 1.39/2.48 & 1.82/3.18 & 1.99/3.84 & 1.51/2.97 & 1.54/2.58 & 1.90/3.25 & 1.32/2.27 & 1.21/2.15 & \underline{1.14}/\underline{1.92} & \textbf{0.87}/\textbf{1.50} \\
\hline
\end{tabular}
\label{table:nfl}}
\vspace{-3mm}
\end{table*}

Technically, we can also explicitly supervise the estimation of leapfrog initializer in stage two, since the distribution $\mathcal{P}(\widehat{\mathbf{Y}}^\tau)$ can be denoised from a normal distribution. For the explicit supervision, we draw $M\gg K$ samples from $\mathcal{P}(\widehat{\mathbf{Y}}^\Gamma)$ under the normal distribution and iteratively denoise these samples through Step~\eqref{eq:denoising_step} until we get expected denoised trajectories $\widehat{\mathbf{Y}}^\tau$. And then, we calculate the statistics of the denoised distribution $\mathcal{P}(\widehat{\mathbf{Y}}^\tau)$ using these $M$ samples, serving as explicit supervisions for mean estimation $f_\mu(\cdot)$ and variance estimation $f_\sigma(\cdot)$. However, since $\tau \ll \Gamma$, we need to run $(\Gamma-\tau)\approx \Gamma$-steps denoising for $M\gg K$ samples to get statistics, resulting in unacceptable time and storage consumption for training (e.g. $\sim$ 6 days per epoch on NBA dataset). We thus do not use explicit supervision.

\subsection{Inference Phase}
During the inference, instead of the $\Gamma$-steps' denoising, leapfrog diffusion model only takes $\tau$-steps, accelerating the inference. To be specific, we first generate $K$ correlated samples to model the distribution $\mathcal{P}(\widehat{\mathbf{Y}}^\tau)$ using the trained leapfrog initializer. Then, these samples will be fed into the denoising process and iteratively fine-tuned to produce the final predictions; see Algorithm \ref{alg:sampling}.

\section{Experiments}
\subsection{Datasets}
We evaluate our method on four trajectory prediction datasets, including two sports datasets (NBA SportVU Dataset, NFL Football Dataset) and two pedestrian datasets (Stanford Drone Dataset, ETH-UCY).

\textbf{NBA SportVU Dataset (NBA)}: NBA trajectory dataset is collected by NBA using the SportVU tracking system, which records the trajectories of the 10 players and the ball in real basketball games. In this task, we predict the future 4.0s (20 frames) using the 2.0s (10 frames) past trajectory.

\textbf{NFL Football Dataset (NFL)}: NFL Football Dataset records the position of every player on the field during each play in the 2017 year. We predict the 22 players’ (11 players per team) and the ball’s future 3.2s (16 frames) trajectory using the historical 1.6s (8 frames) trajectory.

\textbf{Stanford Drone Dataset (SDD)}: SDD is a large-scale pedestrian dataset collected from a university campus in bird's eye view. Following previous works \cite{mangalam2020pecnet, chenxin2022memonet}, we use the standard train-test split and predict the future 4.8s (12 frames) using 3.2s (8 frames) past.

\textbf{ETH-UCY}: ETH-UCY dataset contains 5 subsets: ETH, HOTEL, UNIV, ZARA1, and ZARA2, containing various motion scenes. We use same segment length of 8s as SDD following previous works \cite{mangalam2020pecnet, hu2020collaborative} and use the leave-one-out approach with four sets for training and a left set for testing.

\begin{table*}[!t]
\footnotesize
\centering
\vspace{-13pt}
\setlength{\tabcolsep}{1mm}{\caption{\small Comparison with baseline models on SDD dataset. minADE$_{20}$/minFDE$_{20}$ (meters) are reported. \textbf{Bold}/\underline{underlined} fonts represent the best/second-best result. Our method achieves the best performance in ADE/FDE. $^\ast$ represents the reproduced results from open source.}
\vspace{-10pt}
\fontsize{7.8}{9.4}\selectfont
\begin{tabular}{l|cccccccccc|c}
\hline
\hline
    \multirow{2}{*}{Time} 
    &\makecell[c]{Social-\\GAN\cite{gupta2018social}}
    &\makecell[c]{SOPHIE \cite{sadeghian2019sophie}}
    &\makecell[c]{Trajectron++\\ \cite{salzmann2020trajectron++}} 
    & \makecell[c]{NMMP\cite{hu2020collaborative}}
    & \makecell[c]{Evolve-\\Graph \cite{li2020evolvegraph}}
    & \makecell[c]{PECNet\cite{mangalam2020pecnet}}
    & \makecell[c]{MemoNet\\ \cite{chenxin2022memonet}} 
    &\makecell[c]{NPSN \cite{inhwan2022npsn}}  
    &\makecell[c]{GroupNet\\ \cite{chenxin2022groupnet}}
    &\makecell[c]{MID$^\ast$ \cite{tianpei2022mid}} 
    & \multirow{2}{*}{\textbf{Ours}}
    \\
    & \color{blue}{\scriptsize{CVPR'18}} 
    & \color{blue}{\scriptsize{CVPR'19}} 
    & \color{blue}{\scriptsize{ECCV'20}} 
    & \color{blue}{\scriptsize{CVPR'20}} 
    & \color{blue}{\scriptsize{NIPS'20}} 
    & \color{blue}{\scriptsize{ECCV'20}} 
    & \color{blue}{\scriptsize{CVPR'22}} 
    & \color{blue}{\scriptsize{CVPR'22}} 
    & \color{blue}{\scriptsize{CVPR'22}} 
    & \color{blue}{\scriptsize{CVPR'22}} 
    &
    \\
\hline
\rule{0pt}{9pt}
     4.8s & 27.23/41.44 & 16.27/29.38 & 19.30/32.70 & 14.67/26.72 & 13.90/22.90 & 9.96/15.88 & \underline{8.56}/12.66 & \underline{8.56}/\underline{11.85} & 9.31/16.11 & 9.73/15.32 &  \textbf{8.48}/\textbf{11.66}\\
\hline
\end{tabular}
\label{table:sdd}}
\vspace{-3mm}
\end{table*}

\begin{table*}[!t]
\footnotesize
\centering
\setlength{\tabcolsep}{1mm}{\caption{\small Comparison with baseline models on ETH-UCY dataset. minADE$_{20}$/minFDE$_{20}$ (meters) are reported. \textbf{Bold}/\underline{underlined} fonts represent the best/second-best result. In most subsets, our method achieves the best or second-best performance in ADE/FDE.}
\vspace{-10pt}
\fontsize{7.8}{9.8}\selectfont
\begin{tabular}{l|cccccccccc|c}
\hline
\hline
    
    \multirow{2}{*}{Subset} 
    &\makecell[c]{Social-\\GAN\cite{gupta2018social}}
    &\makecell[c]{NMMP\cite{hu2020collaborative}}
    &\makecell[c]{STAR\cite{yu2020spatio}}
    &\makecell[c]{PECNet\cite{mangalam2020pecnet}}
    &\makecell[c]{Trajectron++\\\cite{salzmann2020trajectron++}}
    & \makecell[c]{Agentformer\\\cite{yuan2021agentformer}}
    &\makecell[c]{MemoNet\\\cite{chenxin2022memonet}}
    & \makecell[c]{$~$NPSN\cite{inhwan2022npsn}$~$}
    &\makecell[c]{GroupNet\\\cite{chenxin2022groupnet}}
    & \makecell[c]{$~$MID\cite{tianpei2022mid}$~$}
    &    \multirow{2}{*}{\textbf{Ours}}
    \\
    & \color{blue}{\scriptsize{CVPR'18}} 
    & \color{blue}{\scriptsize{CVPR'20}} 
    & \color{blue}{\scriptsize{ECCV'20}} 
    & \color{blue}{\scriptsize{ECCV'20}} 
    & \color{blue}{\scriptsize{ECCV'20}} 
    & \color{blue}{\scriptsize{ICCV'21}} 
    & \color{blue}{\scriptsize{CVPR'22}} 
    & \color{blue}{\scriptsize{CVPR'22}} 
    & \color{blue}{\scriptsize{CVPR'22}} 
    & \color{blue}{\scriptsize{CVPR'22}} 
    &
    \\
\hline
ETH   & 0.87/1.62 & 0.61/1.08 & 0.36/0.65 & 0.54/0.87 & 0.61/1.02 & 0.45/0.75 & 0.40/\underline{0.61} & 0.40/0.76 & 0.46/0.73 & \textbf{0.39}/0.66 & \textbf{0.39}/\textbf{0.58} \\
Hotel & 0.67/1.37 & 0.33/0.63 & 0.17/0.36 & 0.18/0.24 & 0.19/0.28 & 0.14/0.22 & \textbf{0.11}/\textbf{0.17} & 0.12/0.18 & 0.15/0.25 & 0.13/0.22 & \textbf{0.11}/\textbf{0.17} \\
Univ  & 0.76/1.52 & 0.52/1.11 & 0.31/0.62 & 0.35/0.60 & 0.30/0.54 & 0.25/0.45 & 0.24/0.43 & \textbf{0.22}/\textbf{0.41} & 0.26/0.49 & \textbf{0.22}/0.45 & 0.26/\underline{0.43} \\
Zara1 & 0.35/0.68 & 0.32/0.66 & 0.29/0.52 & 0.22/0.39 & 0.24/0.42 & 0.18/0.30 & 0.18/0.32 & \textbf{0.17}/\underline{0.31} & 0.21/0.39 & \textbf{0.17}/0.30 & 0.18/\textbf{0.26} \\
Zara2 & 0.42/0.84 & 0.43/0.85 & 0.22/0.46 & 0.17/0.30 & 0.18/0.32 & 0.14/0.24 & 0.14/0.24 & \textbf{0.12}/0.24 & 0.17/0.33 & 0.13/0.27 & \underline{0.13}/\textbf{0.22} \\
\hline
AVG   & 0.61/1.21 & 0.41/0.82 & 0.26/0.53 & 0.29/0.48 & 0.30/0.51 & 0.23/0.39 & \textbf{0.21}/\underline{0.35} & \textbf{0.21}/0.38 & 0.25/0.44 & \textbf{0.21}/0.38 & \textbf{0.21}/\textbf{0.33} \\
\hline
\end{tabular}
\label{table:eth-ucy}}
\vspace{-4mm}
\end{table*}


\begin{figure*}[t]
  \centering
    \vspace{-13pt}
   \includegraphics[width=0.98\linewidth]{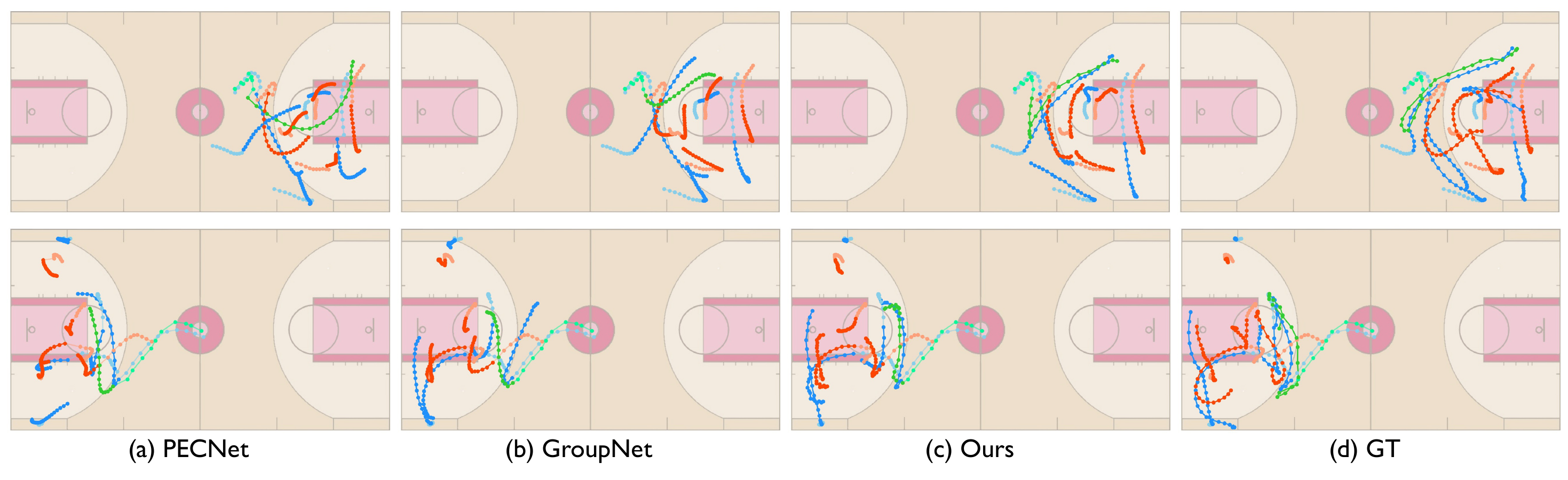}
\vspace{-4.5mm}
   \caption{Visualization comparison on NBA. We compare the best-of-20 predictions by our method and two previous methods. Our method generates a more precise trajectory prediction. (Light color: past trajectory; blue/red/green color: two teams and the basketball.)}
\vspace{-5.5mm}
   \label{fig:performance_comparison}
\end{figure*}

\subsection{Implementation Details}

In the leapfrog diffusion model, we set the diffusion step $\Gamma=100$ for all four datasets and the leapfrog step $\tau=5$ on the NBA dataset. In the leapfrog initializer, we build a transformer-based social encoder where the feed-forward dimension is set to 256, the number of heads is 2, and 2 encoder layers are applied; we apply the temporal encoder with 1D convolution kernel being 3, and output channel setting to 32, and we also build a GRU with the hidden size of 256. In the denoising module, we apply the same parameters transformer to extract the context information, and we build the core denoising module with a hidden size of 256. To train the leapfrog diffusion model, we train the denoising module for 100 epochs with an initial learning rate of $10^{-2}$ and decay to half every 16 epochs. With a frozen denoising module, we then train the leapfrog initializer for 200 epochs with an initial learning rate of $10^{-4}$, decaying by 0.9 every 32 epochs. We set weight parameter $w_1=50$ to emphasize the distance loss. The entire framework is trained with the Adam optimizer on one GTX-3090 GPU. All models are implemented with PyTorch 1.7.1. See more details in the supplementary material.


\begin{table}[!t]
\footnotesize
\centering
\setlength{\tabcolsep}{1mm}{
\caption{Ablation of leapfrog initializer in the leapfrog diffusion model on NFL with various prediction numbers $K$. Each module in the leapfrog initializer is beneficial.}
\vspace{-10pt}
\begin{tabular}{ccc|c|c}
\hline
\hline
\makecell[c]{Mean\\ $\mu_\theta$} & \makecell[c]{Variance\\ $\sigma_\theta$} & \makecell[c]{Sample\\ $\widehat{\mathbb{S}}_{\theta}$} & $K$=2 & $K$=4 \\
\hline
\checkmark  &               & correlated     & 2.04{\tiny{$\pm$0.18}}/4.08{\tiny{$\pm$0.48}} & 1.63{\tiny{$\pm$0.13}}/3.05{\tiny{$\pm$0.16}} \\
            & \checkmark    & correlated     & 1.95{\tiny{$\pm$0.08}}/3.90{\tiny{$\pm$0.22}} & 1.49{\tiny{$\pm$0.01}}/2.86{\tiny{$\pm$0.02}} \\
\checkmark  & \checkmark    & i.i.d     & 2.36{\tiny{$\pm$0.13}}/4.31{\tiny{$\pm$0.22}} & 1.90{\tiny{$\pm$0.07}}/3.31{\tiny{$\pm$0.05}} \\
\checkmark  & \checkmark    & correlated     & \textbf{1.84}{\tiny{$\pm$0.05}}/\textbf{3.61}{\tiny{$\pm$0.11}} & \textbf{1.47}{\tiny{$\pm$0.01}}/\textbf{2.83}{\tiny{$\pm$0.02}} \\
\hline
\makecell[c]{Mean\\ $\mu_\theta$} & \makecell[c]{Variance\\ $\sigma_\theta$} & \makecell[c]{Sample\\ $\widehat{\mathbb{S}}_{\theta}$} & $K$=8 & $K$=20 \\
\hline
\checkmark  &               & correlated     & 1.25{\tiny{$\pm$0.02}}/2.31{\tiny{$\pm$0.04}} & 0.99{\tiny{$\pm$0.03}}/1.68{\tiny{$\pm$0.04}}\\
            & \checkmark    & correlated     & 1.23{\tiny{$\pm$0.01}}/2.20{\tiny{$\pm$0.01}} & 0.95{\tiny{$\pm$0.01}}/1.54{\tiny{$\pm$0.02}} \\
\checkmark  & \checkmark    & i.i.d     & 1.51{\tiny{$\pm$0.04}}/2.67{\tiny{$\pm$0.07}} & 1.18{\tiny{$\pm$0.02}}/1.90{\tiny{$\pm$0.03}}\\
\checkmark  & \checkmark    & correlated     & \textbf{1.18}{\tiny{$\pm$0.01}}/\textbf{2.19}{\tiny{$\pm$0.01}}  & \textbf{0.89}{\tiny{$\pm$0.01}}/\textbf{1.51}{\tiny{$\pm$0.02}} \\
\hline
\hline
\end{tabular}
\label{table:ablation_leapfrog_initializer}}
\vspace{-2.5mm}
\end{table}

\subsection{Comparison with SOTA Methods}

We measure the performance of different trajectory prediction methods using two metrics: minADE$_K$ and minFDE$_K$, following previous work\cite{mangalam2020pecnet, chenxin2022groupnet}. 1) minADE$_K$ calculates the minimum time-averaged distance among $K$ predictions and the ground-truth future trajectory; 2) minFDE$_K$ measures the minimum distance among the $K$ predicted endpoints and the ground-truth endpoints. We calculate these two metrics at different timestamps on sports datasets to better evaluate the performance.

\textbf{NBA dataset.} We compare our method with the current 10 state-of-the-art prediction methods at different timestamps; see Table~\ref{table:nba}. We see that i) our method significantly outperforms all baselines in ADE and FDE at all timestamps. Our method reduces the ADE/FDE at 4.0s from 0.96/1.27 to 0.81/1.10 compared to the current state-of-the-art methods, MID, achieving 15.6\%/13.4\% improvement; and ii) performance improvement over previous methods increases with timestamps, reflecting the proposed method can capture more sophisticated distributions at further timestamps.

\textbf{NFL dataset.} We compare our method with the current 10 state-of-the-art prediction methods at different timestamps; see Table~\ref{table:nfl}. We see that our model significantly outperforms all baselines in ADE and FDE at all timestamps. Our method reduces the ADE/FDE at 3.2s from 1.14/1.92 to 0.87/1.50 compared to the current state-of-the-art methods, MID, achieving 23.7\%/21.9\% improvement. 

\textbf{SDD dataset.} We compare our method with the current 10 state-of-the-art prediction methods; see Table~\ref{table:sdd}. We see that our method reduces FDE from 11.85 to 11.66 compared to the current state-of-the-art method, NPSN. Notably, the original MID~\cite{tianpei2022mid} uses a different protocol from all the other methods, we update its code for a fair comparison.

\textbf{ETH-UCY dataset.} We compare our method with 10 state-of-the-art prediction methods; see Table~\ref{table:eth-ucy}. We see that i) our method reduces FDE from 0.35 to 0.33 compared to the current state-of-the-art method, MemoNet, achieving a 5.7$\%$ improvement; and ii) our method achieves the best or second best to the best performance on most of the subsets.

\begin{table}[!t]
\footnotesize
\centering
\setlength{\tabcolsep}{1mm}{
\caption{Different steps $\Gamma/\tau$ in the standard/leapfrog diffusion model on NBA. $\tau=5$ provides the best performance.}
\vspace{-10pt}
\begin{tabular}{l|l|cccc|c}
\hline
\hline
    Method & Steps & 1.0s & 2.0s & 3.0s & Total(4.0s) & \makecell[c]{Inference\\(ms)}\\
\hline
    \multirow{5}{*}{\makecell[c]{Standard\\Diffusion\\$(\Gamma)$}} & 10   & 0.45/0.51 & 0.98/1.55 & 1.62/2.56 & 2.21/2.77 & $\sim$87\\
                           & 50   &  0.26/0.36 & 0.56/0.91 & 0.89/1.42 & 1.21/1.73 & $\sim$446\\
                           & 100  & 0.21/0.28 & 0.44/0.64 & 0.69/0.95 & 0.94/1.21 & $\sim$886\\
                           & 200  & 0.21/0.29 & 0.44/0.65 & 0.69/0.97 & 0.94/1.21 & $>$1s\\
                           & 500  & 0.21/0.30 & 0.45/0.68 & 0.70/0.99 & 0.95/1.23 & $>$1s\\
\hline
    \multirow{3}{*}{\makecell[c]{Leapfrog\\Diffusion\\$(\tau)$}} & 3    & 0.20/0.31 & 0.40/0.62 & 0.62/0.88 & 0.84/1.10 & $\sim$30\\
                           & 5   &  0.18/\textbf{0.27} & \textbf{0.37}/\textbf{0.56} & \textbf{0.58}/\textbf{0.84} & \textbf{0.81}/1.10 & $\sim$46\\
                           & 10  &  \textbf{0.17}/0.27 & 0.37/0.58 & 0.59/0.85 & 0.82/\textbf{1.08} & $\sim$89\\
\hline
\hline
\end{tabular}
\label{table:ablation_steps}}
\vspace{-5mm}
\end{table}

\subsection{Ablation Studies}

\textbf{Effect of components in leapfrog initializer.} We explore the effect of three key components in leapfrog initializer, including mean estimation, variance estimation, and sample prediction. Table~\ref{table:ablation_leapfrog_initializer} presents the results with mean and variance based on 5 experimental trials. We see that i) the leapfrog initializer achieves stable results with better performance even when prediction number $K$ is small; and ii) the proposed mean estimation, variance estimation, and sample prediction all contribute to promoting prediction accuracy.

\textbf{Effect of leapfrog step $\tau$.} Table~\ref{table:ablation_steps} reports the influence of different leapfrog steps in LED. We see that i) under similar inference time, our method significantly outperforms the standard diffusion model with better representation ability; ii) when $\tau$ is too small, leapfrog initializer targets to learn more sophisticated distribution, causing worse prediction performance; and iii) when $\tau$ is too large, leapfrog initializer has already captured the denoised distribution, encountering performance bottleneck and wasting inference time.

\begin{table}[!t]
\footnotesize
\centering
\setlength{\tabcolsep}{1mm}{
\caption{Comparison to other fast sampling methods on NBA. $\eta=1$ in DDIM. Our method achieves the best performance.}
\vspace{-10pt}
\begin{tabular}{c|cccc|c}
\hline
\hline
    Method & 1.0s & 2.0s & 3.0s & Total(4.0s) & \makecell[c]{Inference \\(ms)}\\
\hline
    PD (K=1) & 0.20/0.33 & 0.45/0.75 & 0.72/1.13 & 0.98/1.39 & $\sim$ 452\\
    PD (K=2) & 0.21/0.34 & 0.46/0.78 & 0.73/1.15 & 0.98/1.41 &$\sim$230\\
    PD (K=3) & 0.23/0.37 & 0.48/0.79 & 0.73/1.15 & 0.98/1.43 &$\sim$121\\
    PD (K=4) & 0.25/0.38 & 0.50/0.80 & 0.75/1.16 & 0.99/1.44 &$\sim$64\\
\hline
    DDIM (S=2) & 0.20/0.29 & 0.42/0.65 & 0.66/0.96 & 0.91/1.21 & $\sim$530\\
    DDIM (S=10) & 0.22/0.32 & 0.44/0.71 & 0.69/1.04 & 0.93/1.31 & $\sim$107\\
    DDIM (S=20) & 0.24/0.35 & 0.49/0.81 & 0.76/1.21 & 1.02/1.51 & $\sim$54\\
\hline
    \textbf{Ours} & \textbf{0.18}/\textbf{0.27} & \textbf{0.37}/\textbf{0.56} & \textbf{0.58}/\textbf{0.84} & \textbf{0.81}/\textbf{1.10} & $\sim$\textbf{46}\\
\hline
\hline
\end{tabular}
\label{table:ablation_other_fast_sampling}}
\vspace{-3.5mm}
\end{table}

\textbf{Comparison to other fast sampling methods.} Table~\ref{table:ablation_other_fast_sampling} compares the performance of our method and the other two fast sampling methods: PD~\cite{Salimans2022pg} and DDIM~\cite{jiaming2021ddim}. We see that our method significantly outperforms two fast sampling methods under similar inference time since the proposed LED promotes the correlation between predictions.

\begin{figure}[!t]
  \centering
   \includegraphics[width=0.99\linewidth]{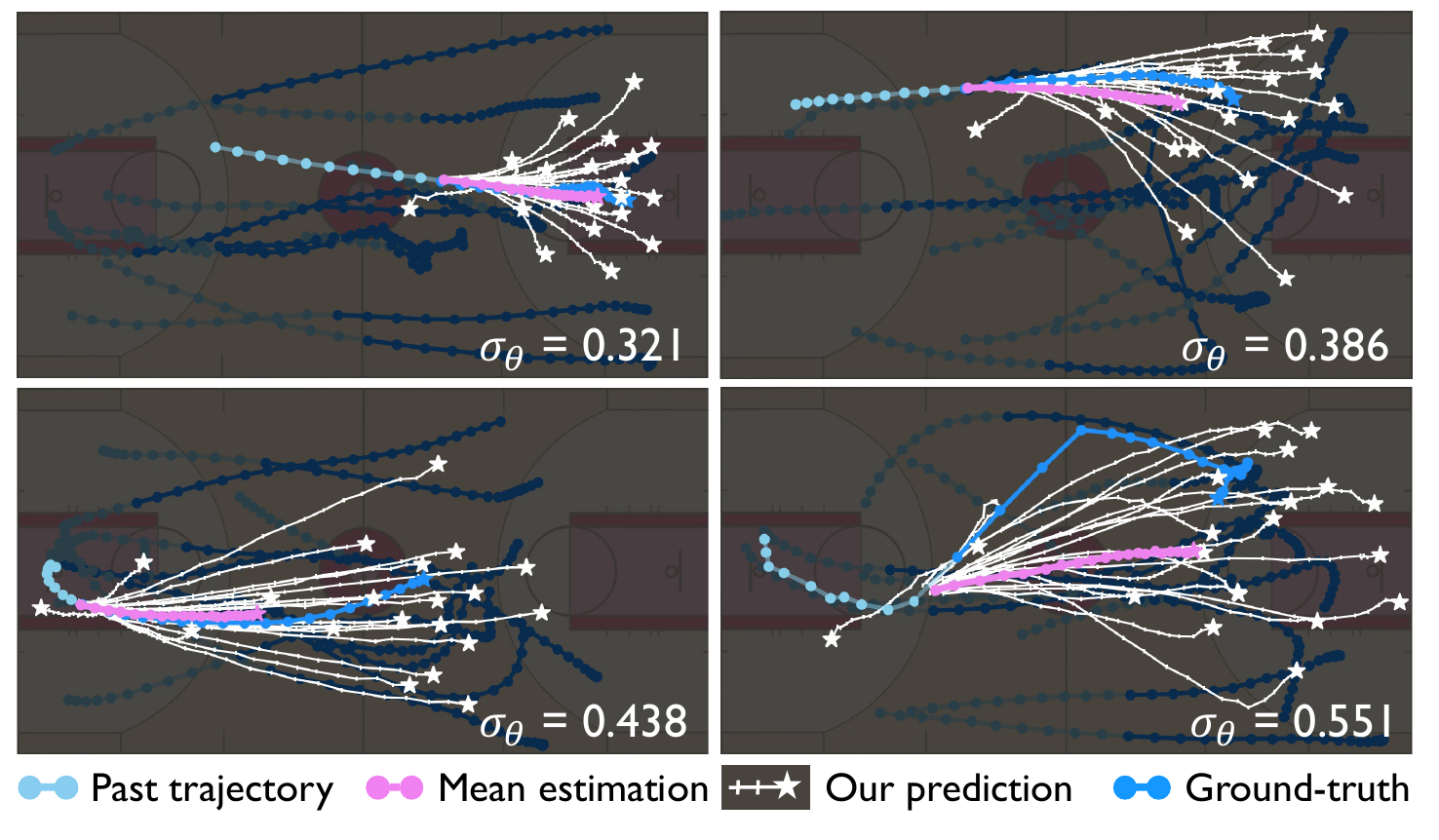}
   \vspace{-3.5mm}
   \caption{Mean and variance estimation in leapfrog initializer on NBA with $K$=20. The estimated variance can reflect the scene complexity of the current agent and produce diverse predictions.}
   \vspace{-6mm}
   \label{fig:visulization_nba}
\end{figure}

\vspace{-1mm}
\subsection{Qualitative Results}
\vspace{-1mm}
\textbf{Visualization of predicted trajectory}. Figure  \ref{fig:performance_comparison} compares the predicted trajectories of two baselines PECNet and GroupNet, our LED (Ours), and the ground-truth (GT) trajectories on the NBA dataset. We see that our method produces more accurate predictions than the previous methods.

\textbf{Visualization of estimated mean and variance}. Figure \ref{fig:visulization_nba} illustrates the mean and variance estimation in the  leapfrog initializer under four scenes on the NBA dataset. We see that the variance estimation can well describe the scene complexity for the current agent by the learned variance, showing the rationality of our variance estimation.

\begin{figure}[!t]
  \centering
   \includegraphics[width=0.99\linewidth]{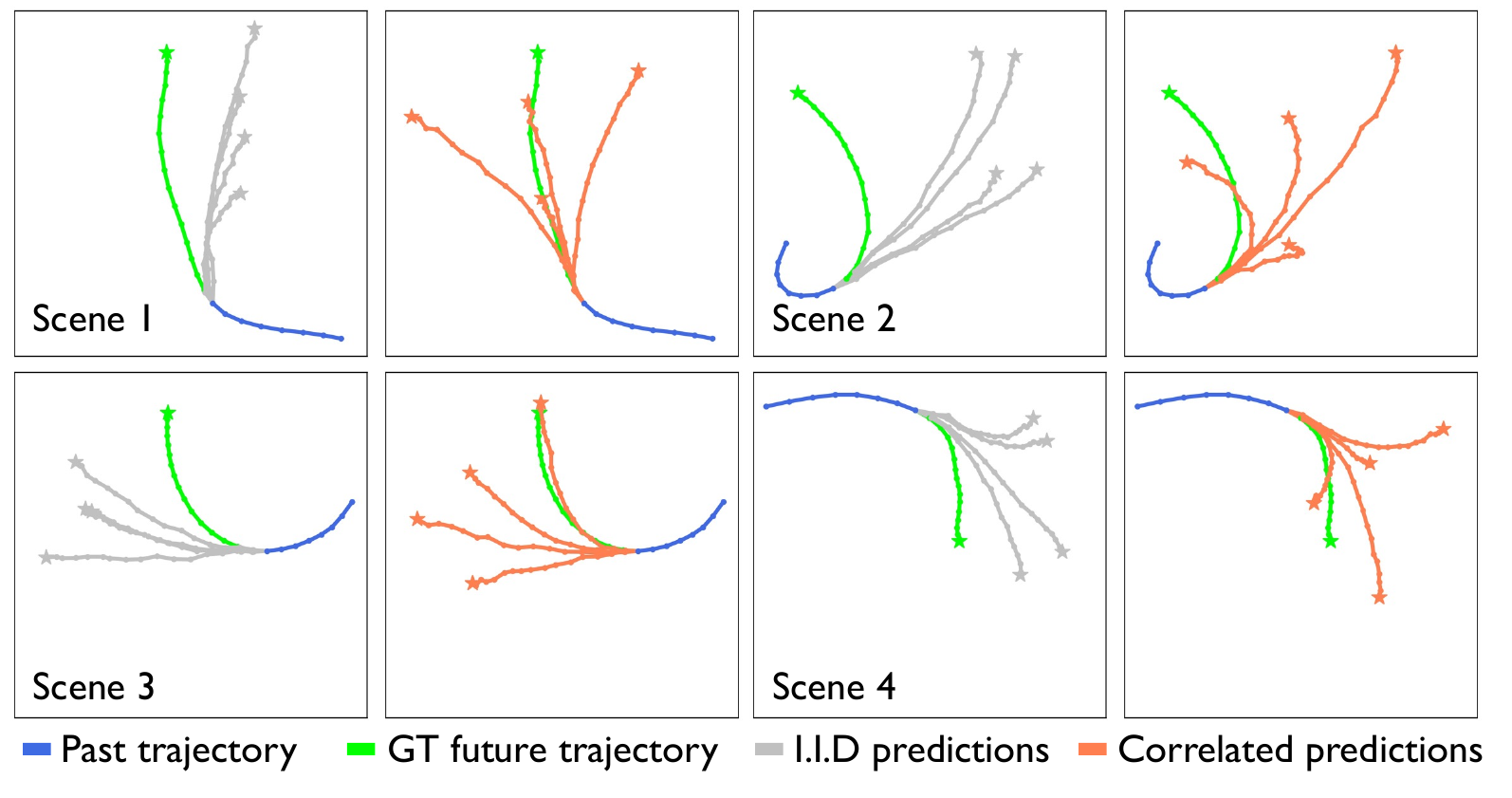}
\vspace{-3.5mm}
   \caption{Comparison between I.I.D and correlated sampling mechanisms in NFL with $K$=4. Correlated samples appropriately capture multi-modalities, significantly improving prediction performances. }
   \label{fig:visulization_sampling}
\vspace{-6mm}
\end{figure}

\textbf{Visualization of different sampling mechanisms}. Figure  \ref{fig:visulization_sampling} compares two sampling mechanisms: I.I.D sampling and correlated sampling in the leapfrog initializer. We see that the proposed correlated sampling can appropriately allocate sample diversity and capture more modalities when the number of trials $K$ is small.

\vspace{-1mm}
\section{Conclusion}
\vspace{-1mm}
This paper proposes the leapfrog diffusion model (LED), a diffusion-based trajectory prediction model, which significantly accelerates the overall inference speed and enables appropriate allocations of multiple correlated predictions. During the inference, LED directly models and samples from the denoised distribution through a novel leapfrog initializer with reparameterization. Extensive experiments show that our method achieves state-of-the-art performance on four real-world datasets and satisfies real-time inference needs.

\textbf{Limitation and future work}. This work achieves inference acceleration for trajectory prediction tasks partially because the dimension of trajectory data is relatively small and the corresponding distribution is much easier to learn compared with those of image/video data. A possible future work is to explore diffusion models and fast sampling methods for higher-dimensional tasks.

\section*{Acknowledgements}
\vspace{-2mm}
This research is partially supported by National Natural Science Foundation of China under Grant 62171276 and the Science and Technology Commission of Shanghai Municipal under Grant 21511100900 and 22DZ2229005.

\clearpage
{\small
\bibliographystyle{ieee_fullname}
\bibliography{ms}
}

\end{document}


\title{Supplementary Material for Leapfrog \\Diffusion Model for Stochastic Trajectory Prediction}

\author{
Chenxin Xu\textsuperscript{1\footnotemark[1]}, Weibo Mao\textsuperscript{1\footnotemark[1]}, Wenjun Zhang\textsuperscript{1}, Siheng Chen\textsuperscript{1,2\footnotemark[2]},
\\\textsuperscript{1}Shanghai Jiao Tong University,  \textsuperscript{2}Shanghai AI Laboratory
\\
\authorcr{\tt\small \{xcxwakaka,kirino.mao,zhangwenjun,sihengc\}@sjtu.edu.cn}
}

\maketitle

\section{Detailed Derivations}

\subsection{Derivation of Standard Diffusion Models}

In the paper submission, we present a standard diffusion model for trajectory prediction following the diffusion-denoising process. Here we elaborate on the details of the diffusion-denoising process.

In standard diffusion models, the diffusion process is operated on the future trajectory $\mathbf{Y}$, while the past trajectories $\mathbf{X}$ and $\mathbb{X}$ serve as a condition for the denoising process. Mathematically, let $\mathbf{Y}^{\gamma}$ be the diffused future trajectory at step $\gamma$, being a basic state in the bidirectional Markov chain of the diffusion-denoising process. We have the start state $\mathbf{Y}^{0} = \mathbf{Y}$ and the end state $\mathbf{Y}^{\Gamma}\sim \mathcal{N}(\mathbf{Y}^{\Gamma}; \mathrm{0}, \mathrm{I})$. We restate the overall procedure of diffusion models for trajectory prediction here, following
\begin{subequations}
\setlength{\abovedisplayskip}{2pt}
   \setlength{\belowdisplayskip}{2pt}
\label{eq:diffusion_model}
    \begin{align}
        \label{eq:initialization_step}
        &\mathbf{Y}^{0}  =  \mathbf{Y},
        \\ 
        \label{eq:diffusion_step}
        &\mathbf{Y}^{\gamma}  =  f_{\operatorname{diffuse}}(\mathbf{Y}^{\gamma-1}),  \ \gamma=1, \cdots, \Gamma,
        \\
        \label{eq:sampling_step}
        &\widehat{\mathbf{Y}}^\Gamma_k  \stackrel{i.i.d}{\sim}   \mathcal{P}(\widehat{\mathbf{Y}}^\Gamma)=\mathcal{N}(\widehat{\mathbf{Y}}^\Gamma; \textbf{0}, \mathbf{I}), {\rm sample}~K~\operatorname{times},
        \\
        \label{eq:denoising_step}
        &\widehat{\mathbf{Y}}^{\gamma}_k  =  f_{\operatorname{denoise}}(\widehat{\mathbf{Y}}^{\gamma+1}_k, \mathbf{X}, \mathbb{X}_{\mathcal{N}}),  \;\gamma\!=\!\Gamma\!-\!1,\!\cdots\!,\!0,
    \end{align}
\end{subequations}
where we use the $f_{\operatorname{diffuse}}(\cdot)$ to represent the diffusion process and $f_{\operatorname{denoise}}(\cdot)$ to represent the conditional denoising process. Here we present the details of these two processes.

\textbf{Forward diffusion process}. Let $(\mathbf{Y}^0, \mathbf{Y}^1, \dots, \mathbf{Y}^\Gamma)$ be the forward $\Gamma$-steps Markov chain constructed by the diffusion model where $\mathbf{Y}^\gamma$ is the diffused future trajectory at step $\gamma$. The forward diffusion process between two steps is defined as
\begin{equation}
\nonumber
    \begin{aligned}
        q(\mathbf{Y}^\gamma|\mathbf{Y}^{\gamma-1}) &= \mathcal{N}(\mathbf{Y}^{\gamma}; \sqrt{1-\beta_\gamma}\mathbf{Y}^{\gamma-1}, \beta_\gamma \mathbf{I}), \\
        \Rightarrow \mathbf{Y}^{\gamma} &= \sqrt{1-\beta_\gamma}\mathbf{Y}^{\gamma-1} + \sqrt{\beta_\gamma} \mathbf{z},
    \end{aligned}
\end{equation}
where $\mathbf{z}\sim \mathcal{N}(\mathbf{z}; 0, \mathbf{I})$ and $\beta_1, \beta_2, \dots, \beta_\Gamma$ are the diffusion parameters controlling the distortion between two steps. In the forward diffusion process, we can directly sample $\gamma$th step diffused trajectory $\mathbf{Y}^{\gamma}$ directly using 
\begin{equation}
\nonumber
    \begin{aligned}
        \mathbf{Y}^{\gamma} &= \sqrt{1-\beta_\gamma}\mathbf{Y}^{\gamma-1} + \sqrt{\beta_\gamma} \mathbf{z}\\
        &\overset{\alpha_\gamma := 1 - \beta_\gamma}{=\!=\!=\!=\!=\!=} \sqrt{\alpha_\gamma} ~\mathbf{Y}^{\gamma-1} + \sqrt{1-\alpha_\gamma} \mathbf{z}\\
        &= \sqrt{\alpha_\gamma} ~(\sqrt{\alpha_{\gamma-1}} ~\mathbf{Y}^{\gamma-2} + \sqrt{1-\alpha_{\gamma-1}} \mathbf{z'}) + \sqrt{1-\alpha_\gamma} \mathbf{z}\\
        &\overset{\operatorname{reparam.}}{=\!=\!=\!=\!=\!=} \sqrt{\alpha_\gamma}\sqrt{\alpha_{\gamma-1}} ~\mathbf{Y}^{\gamma-2} + \sqrt{1-\alpha_\gamma\alpha_{\gamma-1}} \mathbf{z}''\\
        &= \cdots \\
        &= \sqrt{\bar{\alpha}_\gamma}\mathbf{Y}^{0} + \sqrt{1-\bar{\alpha}_\gamma}\mathbf{z},
    \end{aligned}
\end{equation}
where we set the diffusion parameter $\alpha_\gamma := 1 - \beta_\gamma$ and use the reparameterization to merge two Gaussian distributions.
Note that the forward process is non-trainable and with sufficient steps, the final state $\mathbf{Y}^\Gamma \sim q(\mathbf{Y}^\Gamma)$ will be approximate to sample in a normal distribution, i.e., $\mathbf{Y}^\Gamma\sim \mathcal{N}(\mathbf{Y}^\Gamma; \textbf{0}, \textbf{I})$.

\textbf{Conditional denoising process}. Conversely, denote $(\widehat{\mathbf{Y}}^\Gamma, \widehat{\mathbf{Y}}^{\Gamma-1}, \dots, \widehat{\mathbf{Y}}^0)$ as the reverse denoising process conditioned on context information extracting from past trajectories, i.e. $\mathbf{C}=f_{\operatorname{condition}}(\mathbf{X}, \mathbb{X})$. We formulate the conditional denoising process as follows:
\begin{equation}
\label{eq:denoising}
    \begin{aligned}
        & p_\theta(\mathbf{Y}^{\gamma-1} | \mathbf{Y}^{\gamma}, \mathbf{C}) = \mathcal{N}(\mathbf{Y}^{\gamma-1}; \bm{\mu}_\theta^\gamma (\mathbf{Y}^{\gamma}, \mathbf{C}), \beta_\gamma \mathbf{I}), \\
        & \Rightarrow \widehat{\mathbf{Y}}^{\gamma-1} = \bm{\mu}_\theta^\gamma (\widehat{\mathbf{Y}}^{\gamma}, \mathbf{C}) + \sqrt{\beta_\gamma} \mathbf{z},\\
        &\bm{\mu}_\theta^\gamma(\widehat{\mathbf{Y}}^\gamma, \mathbf{C}) = \dfrac{1}{\sqrt{\alpha_\gamma}}(\widehat{\mathbf{Y}}^\gamma - \dfrac{\beta_\gamma}{\sqrt{1-\bar{\alpha}_\gamma}} \bm{\epsilon}_\theta^\gamma(\widehat{\mathbf{Y}}^\gamma, \mathbf{C}))
    \end{aligned}
\end{equation}
where $\mathbf{z}\sim \mathcal{N}(\mathbf{z}; \textbf{0}, \mathbf{I})$, $\alpha_\gamma = 1 - \beta_\gamma$ and $\bar{\alpha}_\gamma = \prod_{\tau=1}^{\gamma} \alpha_\gamma$ are the diffusion parameters at step $\gamma$, and $\bm{\mu}_\theta(\cdot)\in \mathbb{R}^{T_{\rm f}\times 2}$ is the core denoising module with the learnable parameters $\theta$. Note that we have specified the mean term and simplified the variance term in Eq.(\ref{eq:denoising}) following DDPM~\cite{jonathan2020ddpm} so that we can derive the noise estimation loss.

\subsection{Derivation of Noise Estimation Loss}

Here we elaborate on the derivation of our noise estimation loss, the overall target of diffusion models is to maximize the $p_\theta(\mathbf{Y}^{0}|\mathbf{C})$.

\begin{equation}
\nonumber
    \begin{aligned}
     & - \log p_\theta (\mathbf{Y}^{0}|\mathbf{C}) \\
     & \leq -\log p_\theta(\mathbf{Y}^{0}|\mathbf{C}) +\operatorname{D_{KL}} (q(\mathbf{Y}^{1:\Gamma}|\mathbf{Y}^{0}) || p_\theta (\mathbf{Y}^{1:\Gamma}|\mathbf{Y}^{0}, \mathbf{C})) \\ 
     & = -\log p_\theta(\mathbf{Y}^{0}|\mathbf{C}) \\
     & \quad + \int_{\mathbf{Y}^{1:\Gamma}}[\log \dfrac{q(\mathbf{Y}^{1:\Gamma}|\mathbf{Y}^{0})}{p_\theta (\mathbf{Y}^{1:\Gamma}|\mathbf{Y}^{0}, \mathbf{C})}]q(\mathbf{Y}^{1:\Gamma}|\mathbf{Y}^{0})~d\mathbf{Y}^{1:\Gamma} \\
     & = -\log p_\theta(\mathbf{Y}^{0}|\mathbf{C}) + \mathbb{E}_{\mathbf{Y}^{1:\Gamma}\sim q(\mathbf{Y}^{1:\Gamma}|\mathbf{Y}^{0})}[\log \dfrac{q(\mathbf{Y}^{1:\Gamma}|\mathbf{Y}^{0})}{p_\theta (\mathbf{Y}^{1:\Gamma}|\mathbf{Y}^{0}, \mathbf{C})}] \\
     & = -\log p_\theta(\mathbf{Y}^{0}|\mathbf{C}) + \mathbb{E}_{q}[\log \dfrac{q(\mathbf{Y}^{1:\Gamma}|\mathbf{Y}^{0})}{p_\theta (\mathbf{Y}^{0:\Gamma}|\mathbf{C})} + \log p_\theta (\mathbf{Y}^{0}|\mathbf{C})] \\
     & = \mathbb{E}_{q(\mathbf{Y}^{1:\Gamma}|\mathbf{Y}^{0})}[\log \dfrac{q(\mathbf{Y}^{1:\Gamma}|\mathbf{Y}^{0})}{p_\theta (\mathbf{Y}^{0:\Gamma}|\mathbf{C})}] \\
     \end{aligned}
\end{equation}
where we derive the variational lower bound (VLB) to minimize the negative log-likelihood.

\begin{equation}
\nonumber
    \begin{aligned}
     & \Rightarrow  \mathbb{E}_{q(\mathbf{Y}^{0})}- \log p_\theta (\mathbf{Y}^{0}|\mathbf{C}) \leq \mathbb{E}_{q(\mathbf{Y}^{0:\Gamma})}[\log \dfrac{q(\mathbf{Y}^{1:\Gamma}|\mathbf{Y}^{0})}{p_\theta (\mathbf{Y}^{0:\Gamma}|\mathbf{C})}] \\
     &=\mathbb{E}_{q}[\sum_{\gamma=1}^\Gamma - \log \dfrac{p_\theta (\mathbf{Y}^{\gamma-1} | \mathbf{Y}^{\gamma}, \mathbf{C})}{q(\mathbf{Y}^{\gamma}|\mathbf{Y}^{\gamma-1})} - \log p_\theta(\mathbf{Y}^{\Gamma})] \\
     &= - \mathbb{E}_{q}[\sum_{\gamma=2}^\Gamma  \log \dfrac{p_\theta (\mathbf{Y}^{\gamma-1} | \mathbf{Y}^{\gamma}, \mathbf{C})}{q(\mathbf{Y}^{\gamma}|\mathbf{Y}^{\gamma-1})} \\
     & \quad + \log \dfrac{p_\theta (\mathbf{Y}^{0} | \mathbf{Y}^{1}, \mathbf{C})}{q(\mathbf{Y}^{1}|\mathbf{Y}^{0})} + \log p_\theta(\mathbf{Y}^{\Gamma})]\\
     & = - \mathbb{E}_{q}[\sum_{\gamma=2}^\Gamma  \log\Big( \dfrac{p_\theta (\mathbf{Y}^{\gamma-1} | \mathbf{Y}^{\gamma}, \mathbf{C})}{q(\mathbf{Y}^{\gamma-1}|\mathbf{Y}^{\gamma}, \mathbf{Y}^{0})} 
     \cdot \dfrac{q(\mathbf{Y}^{\gamma-1}| \mathbf{Y}^{0})}{q(\mathbf{Y}^{\gamma}| \mathbf{Y}^{0})}\Big) \\
     & \quad +  \log \dfrac{p_\theta (\mathbf{Y}^{0} | \mathbf{Y}^{1}, \mathbf{C})}{q(\mathbf{Y}^{1}|\mathbf{Y}^{0})} + \log p_\theta(\mathbf{Y}^{\Gamma})]\\
     & = - \mathbb{E}_{q}[\sum_{\gamma=2}^\Gamma  \log \dfrac{p_\theta (\mathbf{Y}^{\gamma-1} | \mathbf{Y}^{\gamma}, \mathbf{C})}{q(\mathbf{Y}^{\gamma-1}|\mathbf{Y}^{\gamma}, \mathbf{Y}^{0})} 
     + \sum_{\gamma=2}^\Gamma \log \dfrac{q(\mathbf{Y}^{\gamma-1}| \mathbf{Y}^{0})}{q(\mathbf{Y}^{\gamma}| \mathbf{Y}^{0})} \\
     & \quad + \log  \dfrac{p_\theta (\mathbf{Y}^{0} | \mathbf{Y}^{1}, \mathbf{C})}{q(\mathbf{Y}^{1}|\mathbf{Y}^{0})} + \log p_\theta(\mathbf{Y}^{\Gamma})]\\
     & = -\mathbb{E}_{q}[\dfrac{p_\theta (\mathbf{Y}^{\Gamma})}{q(\mathbf{Y}^{\Gamma}|\mathbf{Y}^{0})} + \sum_{\gamma=2}^\Gamma  \log \dfrac{p_\theta (\mathbf{Y}^{\gamma-1} | \mathbf{Y}^{\gamma}, \mathbf{C})}{q(\mathbf{Y}^{\gamma-1}|\mathbf{Y}^{\gamma}, \mathbf{Y}^{0})}  \\
     & \quad + \log p_\theta (\mathbf{Y}^{0} | \mathbf{Y}^{1}, \mathbf{C})]
     \end{aligned}
\end{equation}
where the first term can be ignored since there are no trainable parameters in $p_\theta (\mathbf{Y}^{\Gamma})$. Then, we only need to focus on the second term $\mathbb{E}_{q}[\log \dfrac{q(\mathbf{Y}^{\gamma-1}|\mathbf{Y}^{\gamma}, \mathbf{Y}^{0})}{p_\theta (\mathbf{Y}^{\gamma-1} | \mathbf{Y}^{\gamma}, \mathbf{C})}]$ where $p_\theta(\cdot)$ is given in Equation~\eqref{eq:denoising}. We can derive the close form for $q(\mathbf{Y}^{\gamma-1}|\mathbf{Y}^{\gamma}, \mathbf{Y}^{0})$ with the Bayes' rule,
\begin{equation}
\nonumber
    \begin{aligned}
        q(\mathbf{Y}^{\gamma-1}|\mathbf{Y}^{\gamma}, \mathbf{Y}^{0}) & = \dfrac{q(\mathbf{Y}^{\gamma-1}, \mathbf{Y}^{\gamma} | \mathbf{Y}^{0})}{q(\mathbf{Y}^{\gamma}| \mathbf{Y}^{0})} \\
        & = \dfrac{q(\mathbf{Y}^{\gamma-1}| \mathbf{Y}^{0})q(\mathbf{Y}^{\gamma} | \mathbf{Y}^{\gamma-1} , \mathbf{Y}^{0})}{q(\mathbf{Y}^{\gamma}| \mathbf{Y}^{0})}\\
        & = \dfrac{q(\mathbf{Y}^{\gamma-1}| \mathbf{Y}^{0})q(\mathbf{Y}^{\gamma} | \mathbf{Y}^{\gamma-1})}{q(\mathbf{Y}^{\gamma}| \mathbf{Y}^{0})}
    \end{aligned}
\end{equation}
where $q(\mathbf{Y}^{\gamma-1}| \mathbf{Y}^{0}), q(\mathbf{Y}^{\gamma} | \mathbf{Y}^{\gamma-1})$, and $q(\mathbf{Y}^{\gamma}| \mathbf{Y}^{0})$ are all Gaussian distributions, which indicates the target distribution $q(\mathbf{Y}^{\gamma-1}|\mathbf{Y}^{\gamma}, \mathbf{Y}^{0})$ also has the Gaussian form. Follow~\cite{jonathan2020ddpm}, suppose $q(\mathbf{Y}^{\gamma-1}|\mathbf{Y}^{\gamma}, \mathbf{Y}^{0}) = \mathcal{N}(\mathbf{Y}^{\gamma-1}; \bm{\mu}^\gamma, \beta_\gamma \mathbf{I})$, where 
\begin{equation}
\nonumber
    \begin{aligned}
        \bm{\mu}^\gamma & = \dfrac{\sqrt{\alpha_\gamma}(1-\bar{\alpha}_{\gamma-1})}{1-\bar{\alpha}_\gamma}\mathbf{Y}^{\gamma} + \dfrac{\sqrt{\bar{\alpha}_{\gamma-1}}\beta_\gamma}{1-\bar{\alpha}_\gamma}\mathbf{Y}^{0} \\
        & = \dfrac{1}{\sqrt{\alpha}_\gamma}(\mathbf{Y}^{\gamma} - \dfrac{\beta_\gamma}{\sqrt{1-\bar{\alpha}_\gamma}}\bm{\epsilon})
    \end{aligned}
\end{equation}
where $\bm{\epsilon}\sim \mathcal{N}(\bm{\epsilon}; \mathbf{0}, \mathbf{I})$. Then, we only need to minimize the means between two distributions and get the noise estimation loss:
\begin{eqnarray}
\nonumber
\label{eq:loss_diffusion}
    \mathcal{L_{\operatorname{NE}}} = \Vert \bm{\epsilon} - \bm{\epsilon}_\theta^\gamma(\widehat{\mathbf{Y}}^\gamma, \mathbf{C})\Vert_2,
\end{eqnarray}


        

\section{Experiment Details}

We apply a standard diffusion model with the diffusion step $\Gamma=100$, the start value $\beta_1=$1e-4, and the end value $\beta_{100}=$5e-2. We use the linear schedule to interpolate the intermediate values $\beta_2, \beta_3, \cdots, \beta_{99}$.

On SDD, following previous destination prediction strategies~\cite{mangalam2020pecnet, chenxin2022memonet}, we first predict the destination of a pedestrian using the proposed leapfrog diffusion model. And then, we fulfill the trajectory using the multi-layer perceptron.

\section{Supplementary Experiments}

\begin{table}[!t]
\footnotesize
\centering
\setlength{\tabcolsep}{1mm}{
\caption{Influence of different parameters in the standard diffusion models on SDD. We run 5 times for each setting with $K$=20 and report the average and best performance.}
\fontsize{8}{11.5}\selectfont
\begin{tabular}{c|c|c|l|ll|ll}
\toprule[1pt]
\multicolumn{4}{c|}{Diffusion Parameters}                                                            & \multicolumn{2}{c|}{AVG}                                 & \multicolumn{2}{c}{Best}                                \\ \hline
$\beta_1$           & $\beta_\Gamma$            & $\Gamma$             & \multicolumn{1}{c|}{schedule} & \multicolumn{1}{c}{minADE} & \multicolumn{1}{c|}{minFDE} & \multicolumn{1}{c}{minADE} & \multicolumn{1}{c}{minFDE} \\ \midrule[0.7pt]
\multirow{8}{*}{1e-4} & \multirow{8}{*}{5e-2} & 20                   & linear                        & 19.27                      & 32.77                      & 10.42                      & 19.19                      \\ \cline{3-8} 
                      &                       & 50                   & linear                        & 11.04                      & 17.75                      & 9.94                      & 15.95                      \\ \cline{3-8} 
                      &                       & \multirow{3}{*}{100} & linear                        & \textbf{10.36 }                     & 16.92                      & \textbf{9.73}                       & \textbf{15.32}                      \\
                      &                       &                      & sigmoid                       & 10.65                      & \textbf{16.87}                      & 9.76                      & 15.52                      \\
                      &                       &                      & quadratic                     & 10.55                      & 17.87                      & 9.84                       & 15.77                      \\ \cline{3-8} 
                      &                       & 200                  & linear                        & 10.70                      & 18.03                      & 10.24                      & 16.98                      \\ \cline{3-8} 
                      &                       & 500                  & linear                        & 10.94                      & 18.68                      & 10.45                      & 17.68                      \\ \cline{3-8} 
                      &                       & 1000                 & linear                        & 11.27                      & 19.01                      & 10.91                      & 18.26                      \\ 
\hline
1e-5                  & 5e-2                  & 100                  & linear                        & 10.43                      & 17.45                      & 9.92                       & 16.03                      \\
\hline
1e-4                  & 1e-2                  & 100                  & linear                        & 25.80                      & 48.29                      & 12.70                      & 21.37                      \\
\hline
1e-5                  & 1e-2                  & 100                  & linear                        & 26.91                      & 44.85                      &  12.52                     & 21.06                      \\
\bottomrule[1pt]
\end{tabular}
\label{table:ablation_parameters}}
\end{table}

\subsection{Influence of Diffusion Parameters}

We explore the influence of different parameters in the diffusion model, including the denoising steps $\Gamma$, the start value of $\beta_1$, the end value of $\beta_\Gamma$, and the schedule to generate $\beta$'s; see Table~\ref{table:ablation_parameters}. We see that i) $\beta_1=1e-4, \beta_\Gamma=5e-2, \Gamma=100$ provides the best performance for the standard diffusion model; ii) with the fixed $\beta_1$ and $\beta_\Gamma$, the schedule to generate the intermediate parameters will not influence the performance lot, also the linear schedule provides the best performance; and iii) when the diffusion step $\Gamma$ is too small, the denoising step is not equivalent to estimating the Gaussian noise, deteriorating the performance.

\begin{table}[!t]
\footnotesize
\centering
\setlength{\tabcolsep}{1mm}{
\caption{Influence of different social-temporal structures in the leapfrog initializer on SDD. We run 5 times for each setting with $K$=20 and report the average and best performance.}
\fontsize{8}{11.5}\selectfont
\begin{tabular}{l|cc|cc}
\toprule[1pt]
\multirow{2}{*}{\begin{tabular}[c]{@{}l@{}}Encoder\\ Structure\end{tabular}} & \multicolumn{2}{c|}{AVG} & \multicolumn{2}{c}{Best} \\ \cline{2-5} 
                                                                             & minADE      & minFDE     & minADE      & minFDE     \\ \midrule[0.7pt]
without social                                                               & 8.69        & 12.07      & 8.64        & 11.93      \\
sequential                                                                   & 8.65        & 11.94      & 8.60        & 11.81      \\
parallel                                                                     & \textbf{8.47}        & \textbf{11.54}      & \textbf{8.46}        & \textbf{11.47}      \\ \bottomrule[1pt]
\end{tabular}
     
\label{table:ablation_encoders}}
\end{table}

\subsection{Influence of Different Encoders}

In the leapfrog initializer, we use a social encoder to capture social influence, a temporal encoder to learn temporal embedding, and an aggregation layer to fuse both social and temporal information. Here we explore the influence of different encoders including without considering the social information (without social), sequential structure to fuse the social-temporal information (sequential), and the parallel structure used in the paper submission (parallel); see Table~\ref{table:ablation_encoders}. We see that i) the parallel structure provides the best performance since the social-temporal information is decoupled without influencing each other; and ii) the social force will influence the agent's movement since considering the social embedding outperforms the without social embedding structure.

{\small
\bibliographystyle{ieee_fullname}
\bibliography{supp}
}